\documentclass[a4paper,fleqn]{cas-sc}

\usepackage[authoryear,longnamesfirst]{natbib}

\usepackage{graphicx}
\graphicspath{ {./figures/} }

\usepackage{caption} 
\usepackage{amsmath,amssymb,amsfonts}

\usepackage[caption=false]{subfig}
\usepackage{diagbox}
\usepackage{comment}
\usepackage{xcolor}
\usepackage[ruled,vlined]{algorithm2e}

\usepackage{adjustbox}
\usepackage[flushleft]{threeparttable}
\usepackage[utf8]{inputenc}
\usepackage{diagbox} 
\usepackage{longtable}
\usepackage{booktabs}
\usepackage{soul}
\usepackage{float}
\usepackage{booktabs, caption, tabularx, makecell}
\usepackage{hyperref}
\usepackage{todonotes}

\def\tsc#1{\csdef{#1}{\textsc{\lowercase{#1}}\xspace}}
\tsc{WGM}
\tsc{QE}
\tsc{EP}
\tsc{PMS}
\tsc{BEC}
\tsc{DE}

\begin{document}
\let\WriteBookmarks\relax
\def\floatpagepagefraction{1}
\def\textpagefraction{.001}

\shorttitle{Tensor-DPMM for Passenger Clustering}

\shortauthors{Ziyue Li et~al.}

\title [mode = title]{Tensor Dirichlet Process Multinomial Mixture Model for Passenger Trajectory Clustering}                      
\tnotemark[1]

\tnotetext[1]{This work is partially funded with the RGC GRF 16201718 and 16216119. The authors appreciate the help from the Hong Kong MTR Co. research, marketing, and customer service teams.}


%
\author[1]{Ziyue Li}[type=author,
                        auid=000,bioid=1,
                        orcid=0000-0003-4983-9352
                        ]

\cormark[1]


\ead{zlibn@wiso.uni-koeln.de}

\ead[url]{https://bonaldli.github.io/}

\credit{Conceptualization of this study, Methodology, Software, Writing - Original draft preparation}

\affiliation[1]{organization={University of Cologne},
    addressline={Albertus-Magnus-Platz}, 
    city={Cologne},
    postcode={50923}, 
    country={Germany}}
\author[2]{Hao Yan}[type=author,
                        auid=000,bioid=1,
                        orcid=0000-0002-4322-7323
                        ]
\fnmark[1]
\ead{hyan46@asu.edu}
\ead[URL]{https://www.public.asu.edu/~hyan46/}
\credit{Conceptualization of this study, Methodology, Writing - Original draft preparation}
\affiliation[2]{organization={Arizona State University},
    city={Tempe},
    postcode={AZ 85281}, 
    country={U.S.A}}
    
\author[3]{Chen Zhang}[%
   orcid=0000-0002-4767-9597
   ]
\fnmark[1]
\ead{zhangchen01@tsinghua.edu.cn}
\ead[URL]{https://thuie-isda.github.io/}

\credit{Conceptualization of this study, Methodology, Writing - Original draft preparation}

\affiliation[3]{organization={Tsinghua University},
    addressline={602 Shunde building}, 
    city={Beijing},
    postcode={100084}, 
    country={China}}

\author[2]{Andi Wang}
\ead{andi.wang@asu.edu}
\ead[URL]{https://web.asu.edu/andi-wang/home}
\credit{Conceptualization of this study, Writing - Original draft preparation}

\author[1]{Wolfgang Ketter}
\ead{ketter@wiso.uni-koeln.de}
\ead[URL]{https://wolfketter.com/}
\credit{Conceptualization of this study}

\author[4]{Lijun Sun}[%
   orcid=0000-0001-9488-0712
   ]
\ead{lijun.sun@mcgill.ca}
\ead[URL]{https://lijunsun.github.io/}

\credit{Conceptualization of this study}

\affiliation[4]{organization={McGill University},
    addressline={817 Sherbrooke St. W.}, 
    city={Montreal},
    postcode={QC H3A 0C3}, 
    country={Canada}}

\author[4]{Fugee Tsung}[%
   orcid=0000-0002-0575-8254
   ]
\ead{season@ust.hk}
\ead[URL]{https://ieda.ust.hk/dfaculty/tsung/}

\credit{Conceptualization of this study, Funding}

\affiliation[5]{organization={The Hong Kong University of Science and Technology},
    addressline={Univresity Road, Clear Water Bay}, 
    city={Kowloon},
    country={Hong Kong}}
\affiliation[6]{organization={HKUST (Guangzhou Campus)},
    city={Guangzhou},
    country={China}}
\cortext[cor1]{Corresponding author}

\fntext[fn1]{Equal Contribution.}

\begin{abstract}
Passenger clustering based on travel records is essential for transportation operators. 
However, existing methods cannot easily cluster the passengers due to the hierarchical structure of the passenger trip information, namely: each passenger has multiple trips, and each trip contains multi-dimensional multi-mode information. 
Furthermore, existing approaches rely on an accurate specification of the clustering number to start, which is difficult when millions of commuters are using the transport systems on a daily basis.

In this paper, we propose a novel Tensor Dirichlet Process Multinomial Mixture model (Tensor-DPMM), which is designed to preserve the multi-mode and hierarchical structure of the multi-dimensional trip information via tensor, and cluster them in a unified one-step manner. The model also has the ability to determine the number of clusters automatically by using the Dirichlet Process to decide the probabilities for a passenger to be either assigned in an existing cluster or to create a new cluster: This allows our model to grow the clusters as needed in a dynamic manner. Finally, existing methods do not consider spatial semantic graphs such as geographical proximity and functional similarity between the locations, which may cause inaccurate clustering. To this end, we further propose a variant of our model, namely the Tensor-DPMM with Graph, where we construct two spatial graphs correspondingly and adopt community detection to learn the graph communities. 

For the algorithm, we propose a tensor Collapsed Gibbs Sampling method, with an innovative step of ``disband and relocating'', which disbands clusters with too small amount of members and relocates them to the remaining clustering. This avoids uncontrollable growing amounts of clusters. A case study based on Hong Kong metro passenger data is conducted to demonstrate the automatic process of learning the number of clusters, and the learned clusters are 4.3 times better measured by within-cluster compactness and cross-cluster separateness. The code will be released in GitHub: \url{https://github.com/bonaldli/TensorDPMM-G}.
\end{abstract}




\maketitle

\section{Introduction}

\begin{figure}[t]
\centering
\includegraphics[width=0.92\columnwidth]{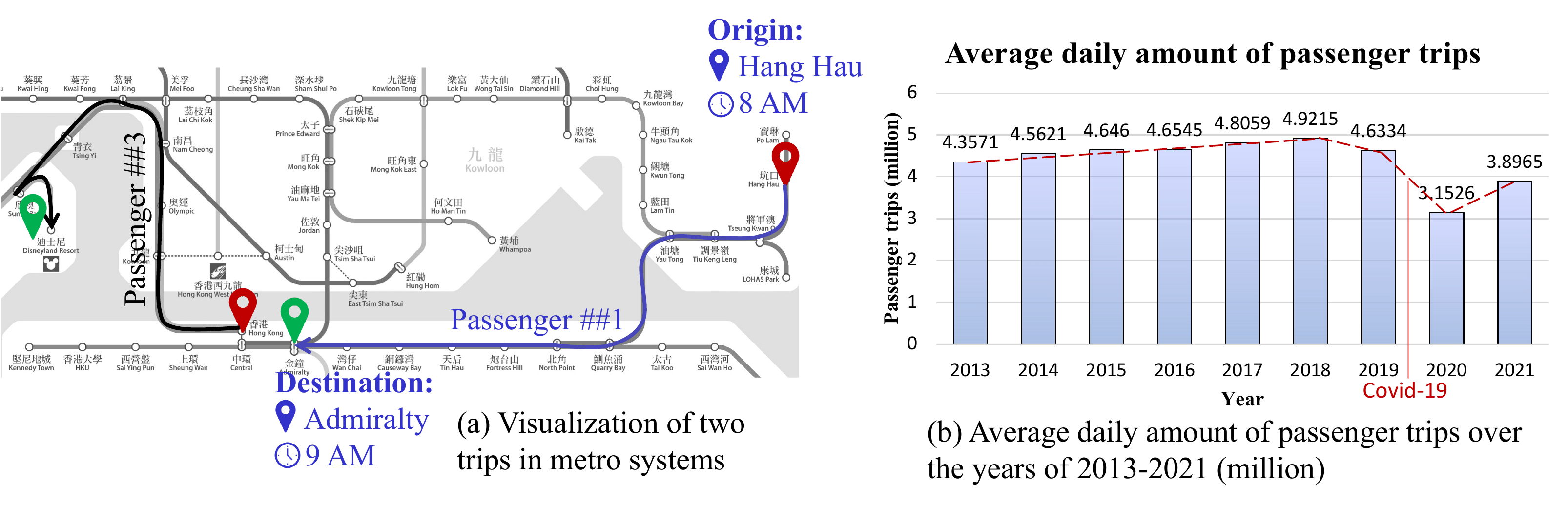}
\caption{(a) Visualization of two trips in metro systems (b) Average daily amount of passenger trips over the years of 2013-2021 (million)}
\label{trip_statistics}
\end{figure}

\begin{figure}[t]
\centering
\includegraphics[width=0.85\columnwidth]{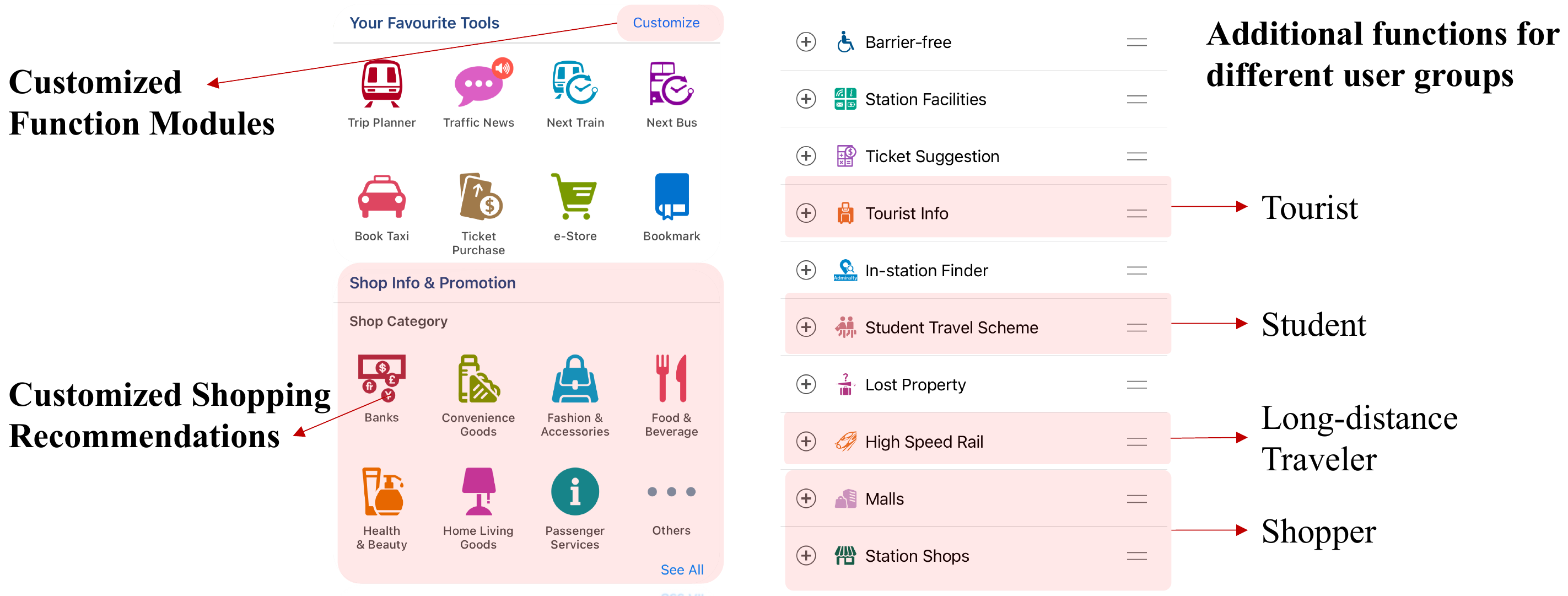}
\caption{The case study of customized services, with customized function modules and customized shopping recommendations}
\label{customized_services}
\end{figure}

Public transportation systems, such as metro networks and bus networks, are important components of urban mobility. As shown in Figure \ref{trip_statistics}.(a), for a public transport system, the passengers' travel record data refer to the trip information of individual passengers, including the passenger's anonymous ID, the origin location, the destination location, departure time, and arrival time for each trip. For example, passenger \#\#1 enters Hang Hau station at 8 AM and exits Admiralty station at 9 AM. These data are typically collected from the Automatic Fare System (AFS) or ticketing machines of the vehicle or station in a large-data scale and real-time manner. As shown in Figure \ref{trip_statistics}.(b), the average daily amount of passenger trips is more than 4 million and constantly increasing before covid-19. In this work, we believe that, based on passengers’ mobility patterns, the service providers could understand the passengers' preferences and profiles better, enabling them to offer more customized services. Passengers' travel records provide rich information on how the passengers commute inside the transport systems, with high potentials of functional and commercial values, such as customized services based on different passenger clustering \citep{ziyue2021tensor}, and other valuable smart mobility functions such as demand prediction (one of the most popular functions) \citep{li2020tensor, 9126133, chen2023adaptive,wang2023correlated,lin2023dynamic}, loading inference \citep{jiang2023unified} (for buses and metros especially), spatial clustering \citep{sun2016understanding}, temporal clustering \citep{cheng2020probabilistic}, travel time estimation \citep{wang2014travel, mao2022jointly}, destination prediction \citep{li2022individualized}, congestion \citep{lan2023mm}, and so on. For example, in Figure \ref{customized_services}, the mobile APP of Hong Kong metro systems could offer different customized function modules for different user groups such as tourists, students, long-distance travelers, shoppers, etc, and under that shopping section, the products and promotions could also be customized.

As an individual-level study of the passengers' travel records, the passenger clustering problem aims to divide the passengers into multiple classes according to the specific characteristics of spatial features such as origin, destination and temporal features such as departure time. The result of the passenger clustering enables the operator to understand the travel patterns of individual groups of passengers and provides commercial values to the system operator by enabling customized operations within each passenger group, such as customized promotions and anomalous trip detection \citep{liu2017personalized,liu2019visual}. Therefore, it is a critical task for achieving the vision of an intelligent transportation system.

\textcolor{black}{To appropriately cluster the passengers into groups, it is essential for the clustering algorithm to fully account for the underlying structure of the passenger record data. 
However, this is challenging because the passenger records data have multitudes of complexities.} 

\begin{figure}[t]
\centering
\includegraphics[width=0.65\columnwidth]{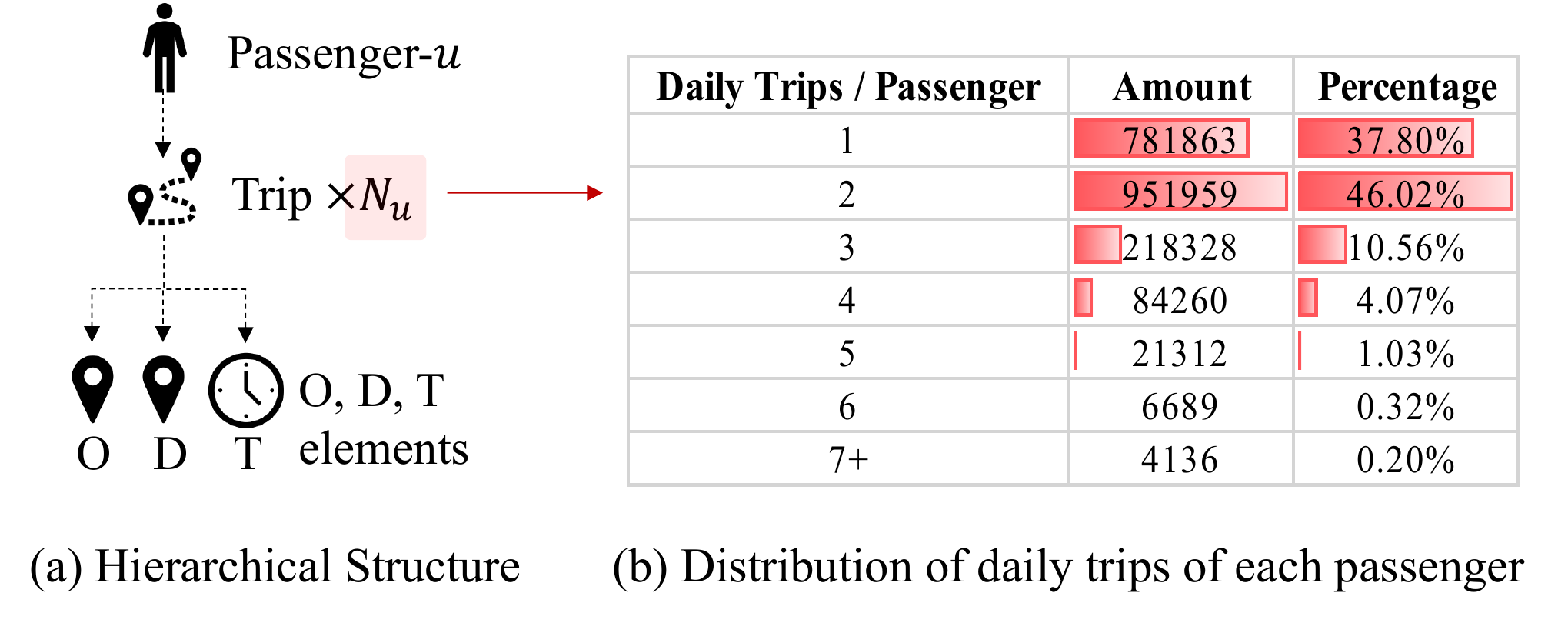}
\caption{(a) Hierarchical structure from passenger to trip, and to ODT elements  (b) Distribution of daily trips of each passenger }
\label{individual_trip_statistics}
\end{figure}

Firstly, we observe that the passenger trip data have a hierarchical structure. As shown in Figure \ref{individual_trip_statistics}. (a), (1) each passenger, e.g., passenger $u$, is associated with the records of multiple trips, e.g., $N_u$ trips, $N_u$ could be different for each passenger, as shown in Figure \ref{individual_trip_statistics}. (b), around 38\% passengers only make one trip per day, and 46\% passengers make two trips per day (mostly back and forth between two stations), and there are a long-tail of more than three trips; (2) under each trip, there are spatiotemporal components including the origin $o$, destination $d$, and the departure time $t$ from fixed ``dictionaries'', namely, there are a fixed amount of stations and time components. 
The clustering algorithm, therefore, should consider the passengers' travel information from both the composition of trips and the spatiotemporal characterization of each trip into account to gain valuable insights about passenger clusters. 

\begin{figure}[t]
\centering
\includegraphics[width=0.7\columnwidth]{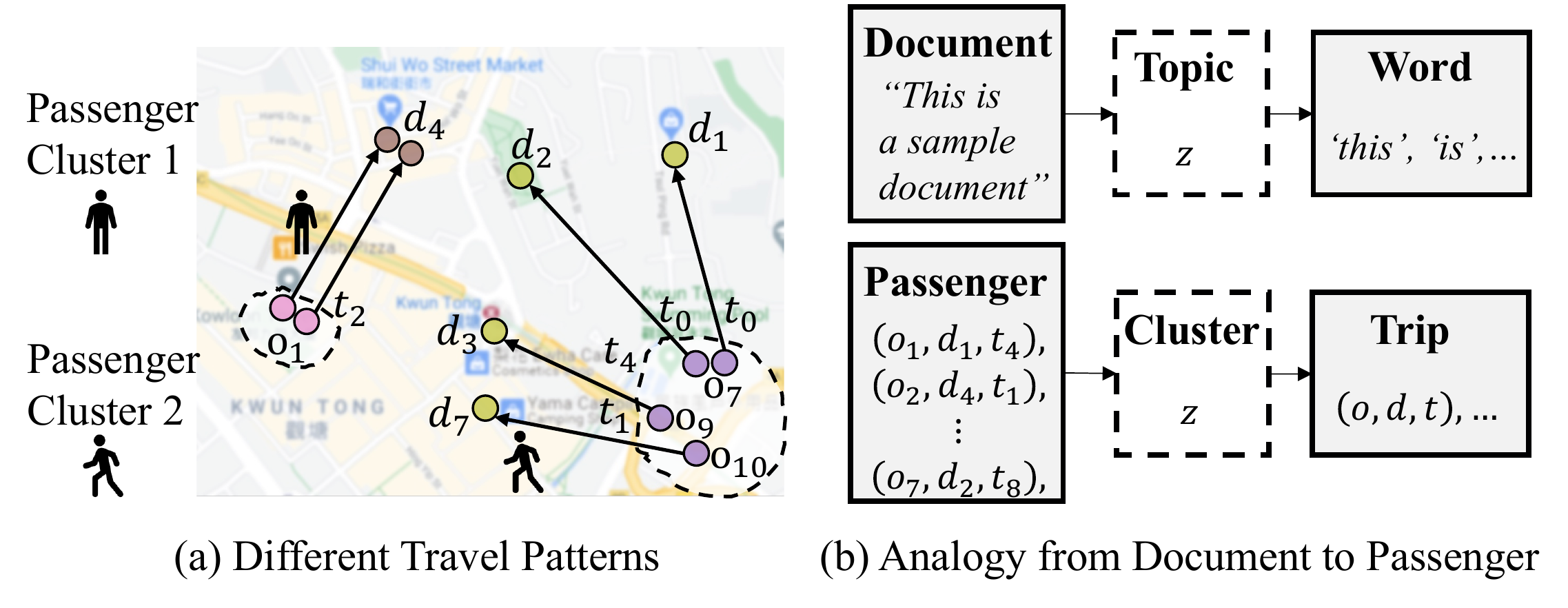}
\caption{(a) Passenger's spatiotemporal trajectories; (b) Analogy from document-topic-word to passenger-cluster-trip}
\label{analogy}
\end{figure}

Secondly, the spatial-temporal structure of passengers within individual clusters is complex and heterogeneous. 
\citep{he2020classification}. 
For instance, as shown in Figure \ref{analogy}. (a), 
passenger cluster 2 has more trips overall, and their trips' spatiotemporal features are more diverse compared to passenger cluster 1, which has a fixed routine. 
Such heterogeneous spatiotemporal structure poses great challenges in data analytics but, on the other hand, is of significant importance in passenger clustering.

\textcolor{black}{Besides the complex structure of passenger travel records, developing passenger clustering methods with low computational complexity is another challenge. In big cities, as mentioned before in Figure \ref{trip_statistics}. (b), the daily trips of public transportation systems are typically in the order of millions and dynamically changing\footnote{\url{https://new.mta.info/coronavirus/ridership}}, which poses great challenges to cluster the huge amount of passengers. As will be discussed in the literature review, a standard two-step distance-based clustering method is to first extract the spatiotemporal features from the passengers and then cluster them by calculating the pairwise distance between their features. However, computing the pairwise distance between passengers involves quadratic complexity and incurs quite a heavy computation burden. As a result, a one-step clustering method without calculating the pairwise distance is especially desired.} 

Finally, the passenger clustering algorithm also requires an automatic decision on the number of clusters. The operators of the transportation systems typically cannot decide the number of passenger clusters because there is a large number of passengers commuting within the transportation system daily, and the constituent of the people is constantly evolving, with new people coming into the system and people quitting using the system over time. Thus, this clustering model should also have the ability to automatically and dynamically determine the number of passenger clusters with the incoming passenger travel records.



In this study, we develop a \textbf{Tensor} \textbf{D}irichlet \textbf{P}rocess \textbf{M}ultinomial \textbf{M}ixture (Tensor-DPMM) model that fully describes the (1) hierarchical structure from passengers to trips to spatiotemporal components, and (2) heterogeneous modes of the complex spatial-temporal structure of passengers within individual clusters. With this model, we propose an algorithm that effectively and efficiently divides all passengers into meaningful clusters.  The algorithm achieves one-step clustering, is computationally efficient without calculating the pairwise distance matrix, and has the capability of automatically determining the number of clusters.  
To the best of our knowledge, this is the first work that fulfills all the three qualities of being spatiotemporal multi-mode, one-step clustering, and enabling the automatic selection of cluster numbers.

Specifically, the Tensor-DPMM model resolve the above-mentioned challenges with the following considerations. (1) To describe the hierarchical relationship between passengers, trips, and the spatiotemporal characteristic of each trip, we innovatively adopt the traditional short text clustering model Dirichlet Multinomial Mixture (\textit{DMM}) \citep{yin2014dirichlet} to the passenger clustering problem. (2) To describe the innate spatiotemporal multi-dimensional data, we propose to adopt tensor formulation \citep{kolda2009tensor}, extending the setting from GR-TensorLDA \citep{ziyue2021tensor}. It allows us to decrease the dimensionality of many origin-destination-time combinations. The integration of the above considerations (1) and (2) enables a one-step clustering approach of the passengers while limiting the computation burden of the method. 
To enable the automatic selection of cluster numbers, we use Dirichlet Process to model the clusters of passengers: we let each passenger either choose an existing cluster or create a new cluster in a flexible, non-parametric, and emerging-with-data manner, saving us from the burden to fix the number of clusters. Although Yin \textit{et al.} \citep{yin2014dirichlet} also extended their DMM approach to Dirichlet Processes (DPMM \citep{yin2016model}), our proposed Tensor-DMM further integrates it with the spatiotemporal tensor model and utilizes the method in passenger clustering application. Last but not the least, \textcolor{black}{in the experiment, we explore the semantic meanings of different stations (e.g., the two stations are geographically close or located in similar functional areas) and proposed an extended version of Tensor-DPMM with semantic graphs considered via graph community detection algorithm.} 

In summary, our technical contributions are as follows: 
\begin{itemize}
   
    \item We developed a tensor DPMM model with the integration of a tensor CP model and an innovative, interdisciplinary application of the traditional short text clustering model \citep{yin2016model} to cluster spatiotemporal passenger trajectory data in a one-step manner with the ability to decide the number of clusters automatically; 

    
    \item We proposed a tensor collapsed Gibbs sampling algorithm and introduced an innovative ``disband and relocate'' step to fulfill a minimal cluster size requirement: if a cluster does not satisfy the minimum size requirement, then it will be disbanded, and all the passengers inside will be relocated to other most similar surviving clusters;
    
    \item As a preliminary solution to incorporate the semantic graphs, we proposed an extension model by pre-processing the stations with graph community detection; 
    
    \item We applied our method to real metro passenger trajectory data based on Hong Kong. We demonstrated that the cluster number could be determined automatically along with the data emerging, and the learned clusters has 4.3 times higher CH-index than benchmark methods, where CH-index measures the inter-cluster separateness over intra-cluster compactness.
    
\end{itemize}

The rest of this paper is organized as follows. In Section  \ref{sec: lr}, we will first briefly review existing passenger clustering methods, including statistical learning-based and deep learning-based methods; Section  \ref{sec: preliminary} will give our model's important preliminaries, including the notation, Dirichlet Multinomial Mixture (DMM) model and Dirichlet Process (DP); Then, in Section  \ref{sec: model}, we propose the first basic model, i.e., Tensor-DMM model, followed by the Tensor-DPMM model with Dirichlet Process. \textcolor{black}{Section \ref{sec: algo_combined} proposes tensor version Gibbs sampling algorithms for Tensor-DPMM and Tensor-DMM, respectively, where in Section \ref{sec: two_prob}, we first derive carefully the probabilities of choosing a cluster that will be used in Gibbs sampling, and} Section  \ref{sec: algo} introduces the algorithm in details; In Section  \ref{sec: experiment}, a real case study is conducted based on Hong Kong Metro data. \textcolor{black}{Moreover, we propose to consider the spatial graphs and further propose our model's final variant, Tensor-DPMM-G model}. Finally, we give the conclusion and future work in Section  \ref{sec: conclude}. The detailed derivation of the two probabilities for the DP is given in Appendix \ref{appx: choose-cluster}.

\section{Literature Review}
\label{sec: lr}
We would like to review state-of-the-art passenger clustering methods. \textcolor{black}{There are usually two types of mainstream methods for clustering passengers' mobility: statistical learning-based methods and deep learning-based methods.}

\subsection{Statistical Methods for Passenger Clustering based on Spatiotemporal Features}


\subsubsection{Temporal Feature Extraction Methods} 
In literature, a class of methods for clustering passengers is based on temporal features such as time-series distance \citep{he2020classification}, active time \citep{briand2017analyzing}, and boarding time \citep{mohamed2016clustering}. \textcolor{black}{For instance, \citet{he2020classification} 
proposed to use Dynamic Time Warping (DTW) \citep{giorgino2009computing} to measure the distance between the time series of AFS transaction data of two passengers \citep{he2020classification}. DTW is known to be insensitive to local compression and stretches and warping, which is a better fit for comparing time series than other distances such as Euclidean distance \citep{deza2009encyclopedia} and Manhattan distance \citep{mori2016distance}. However, this distance method is still a two-step method, with the first step to calculate the distance and the second step to learn the clusters.} Moreover, these methods only consider the features in the temporal dimension yet ignore much information in spatial dimensions. 

\textcolor{black}{\subsubsection{Spatial and Temporal Feature with Two-Step Model} To consider all the spatiotemporal information with a different number of trips, a few models are proposed \citep{cheng2020probabilistic,li2022individualized,yi2019machine,ziyue2021tensor} in a two-step manner: 
\begin{itemize}
    \item In the first step, they learn the hidden features from each dimension with heavy feature engineering burden to extract frequent spatial and temporal features, with the popular spatial features such as co-occurrence, origin-destination pair, transportation mode \citep{yi2019machine}), and temporal features mentioned above. Then,  a passenger could be represented as a probabilistic representation over the extracted features; 
    \item In the second step, passengers are clustered based on their representations using traditional distance-based methods such as K-means or hierarchical clustering methods.
\end{itemize}}

\textcolor{black}{For example, we applied the Latent Dirichlet Allocation (LDA) model to learn the hidden topic of each dimension of ODT and then represented a passenger as a probabilistic distribution over the learned ODT topics \citep{ziyue2021tensor}. Then, in the second step, passengers can be clustered based on their probabilistic representations using traditional distance-based methods. As one of our prior works, this model is named as Graph-Regularized Tensor LDA (GR-TensorLDA) and demonstrates the advantages of applying the traditional topic model in multi-clustering of origin, destination, time, and passenger \citep{li2022individualized}.}

\textcolor{black}{However, these two-step methods suffer from high computational costs and sub-optimal accuracy. For instance, the complexity of calculating the passenger representation's distance is $\mathcal{O}(M^2)$, which is not an optimal choice with millions of the passenger number $M$.}

\textcolor{black}{Besides, another drawback of GR-TensorLDA lies in the basic assumption of LDA-based models, which is: each word has an independent topic assignment. This assumption renders a complex model and a long training time.} 

\textcolor{black}{In this paper, we will borrow the foundational setup from GR-TensorLDA and then highlight the innovation and contribution of this work. Similar to the generative paradigm of GR-TensorLDA, generative models will be preferred to model the hierarchy in the data, i.e., from passenger to a latent cluster, and from  cluster to their belonging trips; Besides, the generative model has the ability to better quantify the uncertainty of the data. Finally, the assumption of the independent topic for each trip will be discarded, which inspires us toward the DMM-based models. More details will be given below as well as in Section \ref{sec: preliminary}.}

\subsubsection{Generative Models as One-Step methods} Recently, generative models have been applied to smart transportation for tasks such as station clustering \citep{mohamed2016clustering}, activity discovery \citep{zhao2020discovering},
destination inference \citep{cheng2020probabilistic}, airline carrier inference \citep{liu2017personalized}, and passenger travel clustering \citep{liu2019visual}. The generative model has the exact power to cluster passengers in a one-step manner, and a few generative models have been proposed, such as uni-gram multinomial mixture \citep{mohamed2016clustering}, Gaussian multinomial mixture \citep{briand2017analyzing}, and Dirichlet multinomial mixture \citep{ziyue2021tensor}.  
The first layer is designed to decide a passenger's membership directly to a cluster. For example, Briand \textit{et al.}, Mahrsi \textit{et al.}, and Li \textit{et al.} all assume that the cluster for a passenger is directly sampled from a multinomial distribution with the number of passenger clusters as $K$. The layers after will be designed to generate other related features that are dependent on the chosen cluster in the first layer.

However, another challenge lies exactly in the lack of prior knowledge to specify the number of clusters. Take the Hong Kong metro system as an example, there are two million passengers daily, and Hong Kong also welcomes thousands of new arrivals flowing into the metro system, rendering it rather difficult for the system operators to specify the number of clusters. To this end, an extension of DMM, namely Dirichlet Process Multinomial Mixture (DPMM), is proposed and also adopted for the first time in this work. It offers the solution of specifying the number of clusters since the $K$ will be automatically increasing when seeing more data.



\subsection{Deep Learning Methods for Passenger Clustering based on Trajectories} 

\textcolor{black}{The common assumption of deep learning-based methods for trajectory modeling is to treat the trajectory as a sequence of the road segments \citep{mao2022jointly}; thus, the popular sequence models from Natural Language Process (NLP), such as Recurrent Neural Network (RNN) \citep{rumelhart1986learning,li2018deep}, Long Short-Term Memory (LSTM) \citep{vaswani2017attention,han2021graph}, and Transformer \citep{vaswani2017attention,lin2021pre} are commonly utilized. For example, the GTS model \citep{han2021graph} first uses Graph Neural Network (GNN) to get the node embedding for each road segment and then inputs the node embeddings into the LSTM to encode the trajectory. Thus, embeddings could be learned for trajectories. Then, a pairwise trajectory similarity could be calculated as the cosine similarity of the embeddings.}

\textcolor{black}{However, this method will not fit in our setting and goal. Firstly, after obtaining the pairwise trajectory similarities, the trajectories could be clustered, yet still in a two-step manner with quadratic complexity with respect to the total number of trajectories, which is even more than the number of passengers since a passenger usually has several trajectories; Secondly, most of the trajectory models are based on the vehicles' trajectories recorded by the Global Positioning System (GPS), for which the entire route locations between origin and destination and their corresponding time stamps are available. However, in our metro system case, only the origin, destination, and corresponding time are available from the AFC data (so our data is not trajectory data, strictly speaking). If there were more route points between OD, the information would still be redundant for a metro system since once given OD information, the travel route between two stations mostly follows the shortest path. Besides, we also prefer a more interpretable, lite, and statistical model for our data, with the ability to determine the number of passenger clusters automatically.}

\section{Preliminary Works}
\label{sec: preliminary}

The method we propose is related to the \textbf{Dirichlet Multinomial Mixture (DMM)} \citep{yin2014dirichlet}, a hierarchical model for short-text clustering, as well as the \textbf{Dirichlet Process (DP)} \citep{ferguson1973bayesian}, which enables the automatic decision of the number of clusters. In this section, we will first summarize the notations and then give the preliminaries about the two basic models.

\subsection{Notations}

Throughout this paper, scalars are denoted in italics, e.g., $n$; vectors by lowercase letters in boldface, e.g., $\mathbf{u}$; matrices by uppercase boldface letters, e.g., $\mathbf{U}$; and sets by either uppercase regular letters $O,D,T$ or boldface script capital $\boldsymbol{\mathcal{D}}$. 
All the notations are summarized in Table \ref{notation}. 

\begin{table}[t]
  \caption{Notation}
  \begin{tabularx}{\linewidth}{lX}
    \toprule
    Symbols & Description \\
    \midrule
    ${()}^{O, D, T}$ & superscripts O, D, T means three dimensions: Origin, Destination, Time respectively\\
    $\mathcal{D}$ & the entire set of the passengers $\mathcal{D} = \{ \mathbf{d}_u\}_{u=1}^{M}$\\
    $M$ & the total number of passengers (iterator $u$)\\
    $\mathbf{d}_u$ & the trip records of the $u$-th passenger $ \mathbf{d}_u = \{\boldsymbol{w}_1, \dots, \boldsymbol{w}_i, \dots, \boldsymbol{w}_{N_u}\}$ \\
    $N_u$ & the number of trips (words) in $\mathbf{d}_u$ (iterator $i$)\\
    $K$ & the number of clusters, and it can be $\infty$ in the Dirichlet process setting\\
    $z_u$ & the cluster assignment for $u$-th passenger $z_u = 1, \dots, K$; \\
    $\mathbf{w}$ & the three-dimensional trip record $\boldsymbol{w} = (w^O, w^D, w^T)$\\
    $w^O_{i}, w^D_{i}, w^T_{i}$ & the origin, destination and time of the $i$-th trip\\
    $V^O, V^D, V^T$ & the vocabulary of all unique origin, destination, and time\\
    $\boldsymbol{\theta}$ & $\boldsymbol{\theta} \in \mathbb{R}^K$ the population-wide Multinomial parameter\\
    $\boldsymbol{\phi}_k^O, \boldsymbol{\phi}_k^D$, $\boldsymbol{\phi}_k^T$ & $\boldsymbol{\phi}_k^O \in \mathbb{R}^{V^O}$ the Multinomial parameter for origins given the cluster $k$, similar for $\boldsymbol{\phi}_k^D$, $\boldsymbol{\phi}_k^T$\\
    $\boldsymbol{\alpha}$ & $\boldsymbol{\alpha} \in \mathbb{R}^{K}$ the Dirichlet prior parameter for $\boldsymbol{\theta}$\\
    $\boldsymbol{\beta}^O, \boldsymbol{\beta}^D$, $\boldsymbol{\beta}^T$ & $\boldsymbol{\beta}^O, \boldsymbol{\beta}^D$, $\boldsymbol{\beta}^T \in \mathbb{R}^{K}$ the Dirichlet prior parameter for $\boldsymbol{\phi}_k^O, \boldsymbol{\phi}_k^D, \boldsymbol{\phi}_k^T$, respectively\\
    $\mathcal{D}_{-u}$ ($\mathcal{D}_{z, -u}$) & all the passengers (in the cluster $z$) except the $u$-th one\\
    $m_{z, -u}$ & the number of passengers in the cluster $z$ except the passenger $u$\\
    $\mathbf{z}_{-u}$ & the cluster assignment vector without the $u$-th passenger\\
    $N^{w}_u$ & the number of occurrences of word $w$ in the $u$-th document regardless ODT \\
    $n_{z,-u}^{w}, n_{z,-u}$ & the number of occurrences of word $w$ in cluster $z$ and the total number of words in cluster $z$ without considering $u$-th document, respectively, and $n_{z,-u} = \sum_{w=1}^{V} n_{z,-u}^{w}$ \\
    $\mathbf{G}_{adj}, \mathbf{G}_{poi}$ & two semantic graphs, i.e., geographical proximity graph and functional similarity graph\\
    $r$ & the minimal cluster size requirement \\
    \bottomrule
  \end{tabularx}
  \label{notation}
\end{table}

\subsection{The Dirichlet Multinomial Model: DMM} 
\label{sec: dmm}
The DMM model \citep{yin2014dirichlet} assumes that a document $\mathbf{d}$ is composed of multiple words and is generated by the following mixture model: 

\begin{itemize}
    \item [(1)] Each document $\mathbf{d}$ may fall into one of $K$ topics, and the topic is firstly selected with weights $P(z=k)$. \textcolor{black}{The corresponding generative process is designed as a multinomial distribution parameterized by $\boldsymbol{\theta}$, thus $P(z=k) = P(z=k|\boldsymbol{\theta}) = \theta_k$. To further design a prior distribution for the multinomial distribution, DMM assigns a Dirichlet prior parameterized by $\boldsymbol{\alpha}$, i.e., $\boldsymbol{\theta} \sim \text{Dir}(\boldsymbol{\alpha})$, where both  $\boldsymbol{\theta}, \boldsymbol{\alpha} \in \mathbb{R}^K$.} 
    \item [(2)] Given the cluster $k$, $\mathbf{d}$ is generated from distribution $p(\mathbf{d} | z=k)$, where all words from this document are generated independently according to the distribution specified by the topic $k$. \textcolor{black}{Thus, we have $p(\mathbf{d} | z=k) = \prod_{ w \in \mathbf{d}} P(w | z=k)$. $P(w | z=k)$, the selection of a word for the given topic $k$, is also designed as a multinomial distribution with parameter $\boldsymbol{\phi}_k$. Similarly, this multinomial parameter is also generated from a Dirichlet prior $\boldsymbol{\beta}$: $\boldsymbol{\phi}_k \sim \text{Dir}(\boldsymbol{\phi})$, where both $\boldsymbol{\phi}_k, \boldsymbol{\beta} \in \mathbb{R}^{V}$, and $V$ is the vocabulary size.}
\end{itemize}

\noindent The likelihood of $\mathbf{d}$ could be represented as a mixture over all the $K$ clusters: 
\[P(\mathbf{d}) = \sum_{k=1}^{K} P(\mathbf{d} | z=k) P(z=k)=\sum_{k=1}^{K} P(z=k)\prod_{ w \in \mathbf{d}} P(w | z=k).\] 

The key implication of the model is that the whole text only has exactly one topic. This assumption is generally satisfied when the length of the text is short (e.g., around $10^2$). 

\textcolor{black}{A Gibbs Sampling algorithm was proposed \citep{yin2014dirichlet}  to solve the DMM, with the probability of the $u$-th document $\mathbf{d}_u$ being assigned to cluster membership $z$ derived as follows:}

\begin{equation}\label{eq_dmm}
\begin{split}
    P(z_u = z | \mathbf{z}_{-u}, \mathcal{D}, \mathbf{\alpha}, \mathbf{\beta}^O, \mathbf{\beta}^D, \mathbf{\beta}^T) \propto \frac{m_{z,-u} + \alpha}{M-1+K\alpha} \frac{\prod_{w \in \mathbf{d}_u} \prod_{j=1}^{N^{w}_u} (n_{z,-u}^{w} + \mathbf{\beta} +j - 1) }{\prod_{i=1}^{N_u}(n_{z,-u} + V \beta + i - 1)}.
\end{split}
\end{equation}

\textcolor{black}{where $K$ is the number of clusters, $M$ is the total number of members, $\alpha, \beta$ are prescribed parameters, $\mathbf{z}_{-u}$ is the cluster assignment vector without the $u$-th member, $m_{k, -u}$ is the number of members in the cluster $k$ except the $u$-th member, and $N^{w}_u$ is the amount of occurences of the word $w$ in the document $\mathbf{d}_u$.}
\subsection{Dirichlet Process (DP)} 

\textcolor{black}{The DMM model could achieve one-step clustering. However, it requires a fixed and finite $K$, which is hard to be defined for dynamic and large-scale public transport systems. With a fixed $K$, the DMM model is thus regarded as a parametric model. Instead, a flexible $K$ is desired, which could emerge with the data. As a result, this flexibility could allow an emergent property in model parameterization, further encouraging the model complexity to be able to grow adaptively with data increasing \citep{li2019tutorial}. Therefore, our model achieves being non-parametric.} 

\begin{figure}[t]
\centering
\includegraphics[width=0.7\columnwidth]{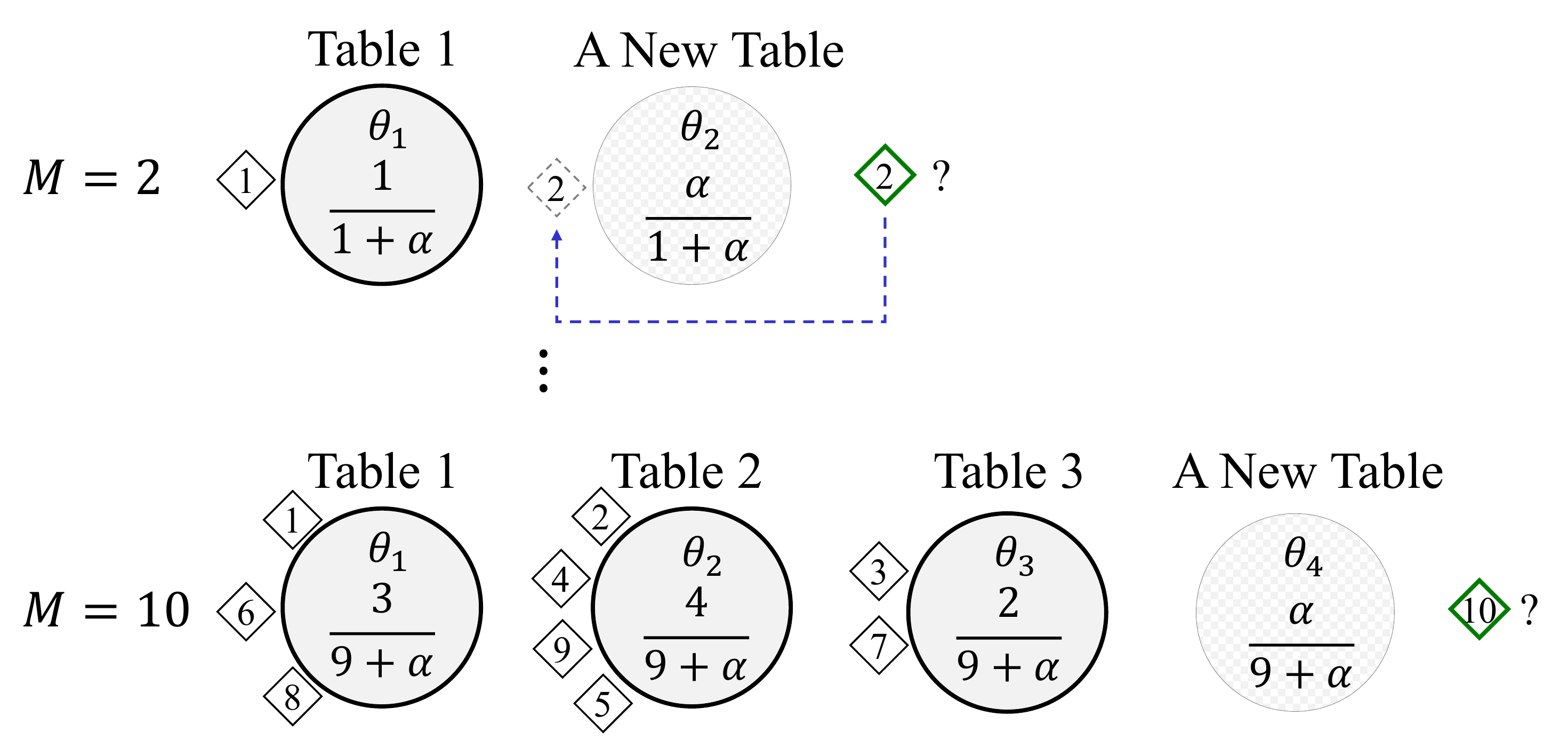}
\caption{Demo of Dirichlet Process (Chinese Restaurant Process): For example, the 10-th passenger chooses a cluster in a DPMM manner, with the possibility of choosing existing clusters related with $m_{z,-u}$, and choosing a new cluster is related with $\alpha$.}
\label{fig: CRP}
\end{figure}

\textcolor{black}{Dirichlet Process (DP) \citep{gershman2012tutorial,li2019tutorial} allows the number of clusters to vary, either being finite or infinite. We follow the same notations and parameter generative processes described in Section \ref{sec: dmm}; thus, the probability of the cluster assignments is derived as below. For more details, please refer to the work \citep{li2019tutorial}:} 

\begin{equation}\label{eq: p_all_z}
\begin{split}
     P(z_1, \dots, z_M) & = \int P(z_1, \dots, z_M  \mid \theta_1, \dots, \theta_M) \times P (\theta_1, \dots, \theta_M \mid \alpha) d\theta_1 \dots d\theta_M \\
     & = \frac{\Gamma(\alpha)}{\Gamma(\alpha/K)^k} \int \prod_{k=1}^{K} \theta^{m_k+ \alpha/K -1} d\theta_k = \frac{\Gamma(\alpha)}{\Gamma(M+\alpha)} \prod_{k=1}^{K} \frac{\Gamma(m_k + \alpha / K)}{\Gamma(\alpha/K)}
\end{split}
\end{equation}
Specifically, based on the Eq. (\ref{eq: p_all_z}), the cluster assignment for the $u$-th passenger given all other assigned clusters except $z_u$ could be derived as the following conditional prior:

\begin{equation}
    P(z_u = k | \mathbf{z}_{-u}, \alpha) = \frac{m_{k, -u}+\alpha/K}{M-1+\alpha}
\end{equation}

As mentioned before, $K$ is not fixed any more in the DP. Flexible $K$ means an infinite number of clusters. We let $K  \to	 \infty$, then we could eliminate the term $\alpha/K$ in the numerator and get the two non-parametric priors for the stochastic process: For a new member $u$, he/she can either choose an existing non-empty cluster or choose a new empty cluster, as shown in Figure \ref{fig: CRP}.

\textbf{When choosing any existing non-empty cluster}: We have $m_{k, -u} > 0$,  then $P(\mathbf{z}_u=k | \mathbf{z}_{-u}, \alpha )$ is obtained as follows:
\begin{equation} \label{eq_prel_choosek}
    P(z_u = k | \mathbf{z}_{-u}, \alpha) = \frac{m_{k, -u}}{M-1+\alpha},
\end{equation}

\textbf{When choosing a new empty cluster}: Given the total probability of having chosen all the existing clusters is $\sum_k P(z_u = k | \mathbf{z}_{-u}, \alpha) = \frac{\sum_k m_{k, -u}}{M-1+\alpha} = \frac{M-1}{M-1+\alpha}$ (the numerator includes all the passengers except the $u$-th one), the probability of choosing a new cluster $z_u=K+1$ can be derived as follows:
\begin{equation} \label{eq_prel_choosenew}
    P(z_u = K+1| \mathbf{z}_{-u}, \alpha) = 1 - \sum_k P(z_u = k | \mathbf{z}_{-u}, \alpha) = \frac{\alpha}{M-1+\alpha}.
\end{equation}


The DMM and DP provide the probabilistic framework for describing the hierarchical nature of the passenger travel records and the flexible clustering. Nevertheless, they are not directly applicable to multi-dimensional data like passenger trajectory data. In the next section, we integrate them with a tensor model that describes the spatiotemporal travel pattern for one-step passenger clustering.

\section{Proposed Models: Tensor Multinomial Mixture (Tensor-DMM) and Tensor Dirichlet Process Multinomial Mixture (Tensor-DPMM)}
\label{sec: model}
\textcolor{black}{In this section, the detailed formulation of the proposed models will be introduced. Section \ref{sec: data_format} will first formulate the data. Section \ref{sec: likelihood} then gives the likelihood formulation based on CP decomposition; Then, in Section \ref{sec: TensorDMM}, we introduce the first proposed Tensor-DMM model, the extended tensor version from the DMM model. However, the Tensor-DMM model has the drawback of relying on a specified and fixed $K$ and lack of flexibility. To solve this, in Section \ref{sec: TensorDPMM-G}, we will introduce the second proposed model, TensorD\textbf{P}MM, which introduces Dirichlet Process. 
Last but not least, the most important step is to \textcolor{black}{ develop the cluster selection probability, similar to Eq. (\ref{eq_dmm}), yet in a tensor version. We start to introduce with the full model: for Tensor-DPMM, we} formulate the probabilities of choosing an existing cluster (in Section \ref{sec: pz}) and creating a new cluster (in Section \ref{sec: pK+1}) \textcolor{black}{correspondingly; quite similarly, for Tensor-DMM, we formulate the probabilities of choosing an cluster from a fixed amount of $K$ clusters} .} 

\subsection{Problem Formulation}
\label{sec: data_format}
In this work, the passenger travel record we are studying has a general format, which commonly contains the information of the passenger's anonymous ID, origin location, destination location, departure time, and arrival time. 
Specifically, we assume that we have the travel records for a total of $M$ passengers, and the passenger $u\in \mathbb{R}^M$ traveled $N_u$ trips. 
The $i$-th trip of passenger $u$ is thus defined as a triplet $\boldsymbol{w}_{u,i} = (w_{u,i}^O, w_{u,i}^D, w_{u,i}^T)$, indicating that this passenger started from origin $w_{u,i}^O\in \mathbb{R}^{V^O}$ within the timeslot $w_{u,i}^T\in \mathbb{R}^{V^T}$, and arrived at destination $w_{u,i}^D\in \mathbb{R}^{V^D}$ for this trip. 
Here $V^O$ and $V^D$ denote the amounts of origins and destinations, respectively. The time intervals within a day are prescribed, and $V^T$ denotes the set of all time intervals. 
All trips from passenger $u$ can thus be represented as $\mathbf{d}_u = \{\boldsymbol{w}_{u,1}, \dots, \boldsymbol{w}_{u, N_u}\}$. 
The entire data are $\boldsymbol{\mathcal{D}} = \{ \mathbf{d}_u\}_{u=1}^{M}$. 
We use the notation $\mathbf{d}_u$ and $\boldsymbol{w}_{u,i}$ because they correspond to the concept of ``document (short-text with small bag-of-words)'' and ``word'' in DMM \citep{yin2014dirichlet}, the short-text classification literature. The analogy is shown in Figure \ref{analogy}.(b) and will be clear after the introduction of the methodology.

\subsection{Likelihood Function for Passenger Travel Record}
\label{sec: likelihood}
Recall that there are two challenges to the structure of the travel record data: (1) the travel records have a hierarchical structure from passengers to trips (2) the trips have heterogeneous modes of the complex spatial-temporal structure. To tackle these challenges, the Tensor-DPMM (1) uses a mixture model to describe the travel records for the passenger $\mathbf{d}_u$. and (2) uses a tensor CP structure to describe the spatial-temporal structure. 

\subsubsection{Mixture model.} We assume that all passengers can be divided into clusters and that each passenger $u$'s travel record $\mathbf{d}_u$ follows a specific spatiotemporal model 
according to the clustered index $z_u$ of this passenger. 
Specifically, the distribution of  $\mathbf{d}_u$ can be represented as follows:
\begin{equation}
    P(\mathbf{d}_u) = \sum_k  P(z_u=k) P(\mathbf{d}_u|z_u=k).
\label{eq_hier}
\end{equation}
Furthermore, we assume that all trips of all passengers within cluster $k$ are  independent, and are associated with a specific travel pattern cluster, as the ``words'' in a ``topic'' follow a specific distribution short-text LDA model as follows:
\begin{equation}
    P(\mathbf{d}_u|z_u=k) = \prod_{i=1}^{N_u} P(\boldsymbol{w}_{u,i} | z_u=k)=\prod_{i=1}^{N_u} P_k(\boldsymbol{w}_{u,i} ).
    \label{eq_shorttxt}
\end{equation}
Here, $P_k(\boldsymbol{w}_{u,i})$ is the probability distribution of a trip $\boldsymbol w$ for a passenger within cluster $k$. 

The model above is motivated by the short-text-based model DMM, which assumes each piece of short text has only one topic, and the probability distribution of each word in the short text is independent and only relies on the topic of this short text. 
The major difference between the short text model and existing LDA-based models \citep{cheng2019inferring,cheng2020probabilistic,ziyue2021tensor,li2022individualized} is that the LDA model assumes that each text has a \textit{mixture} of topics and that each word randomly comes from one of the topics, determining its distribution. Likewise, the Tensor-DPMM model is different from the Tensor LDA models 
\citep{ziyue2021tensor,cheng2020probabilistic}, which assume that each trip has its own ``topic'', and a passenger is a distribution over different ``topics''. 
As we scrutinize the passengers' travel records, we determine that the short-text model is more appropriate for passenger travel record data. 
\begin{itemize}
\item On the one hand, it is observed that 84\% of the passengers only have one or two trips per day, so the average number of trips of a passenger over a whole year is around $10^2$. 
Recall that in short-text classification literature, documents with around $10^2$ words are deemed as a ``short'' text. 
\item On the other hand, we observed that most passengers usually have a fixed travel purpose (school, work, etc.), so trips from all passengers from one cluster should share one spatiotemporal pattern. 
\end{itemize}
Furthermore, our Tensor-DPMM model has one less layer than the LDA model. It is a great benefit as it relieves heavy computation.

\begin{figure}[t]
\centering
\includegraphics[width=0.8\columnwidth]{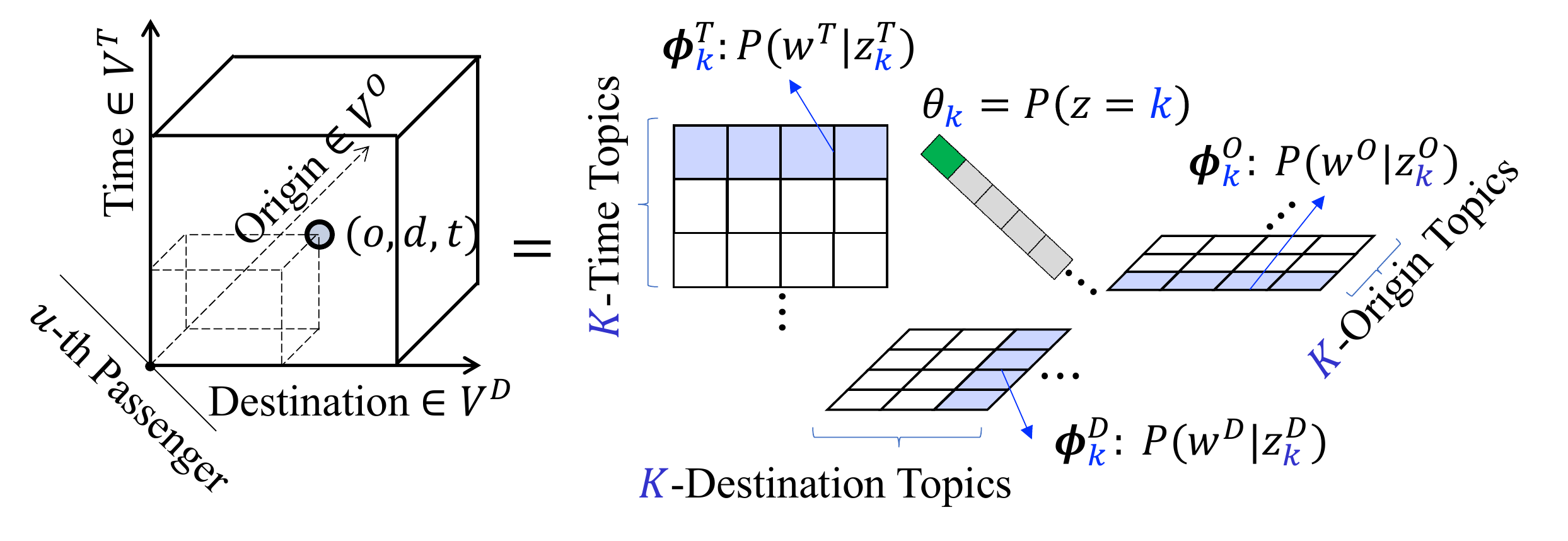}
\caption{The Tensor-DMM and Tensor-DPMM are essentially CP Decomposition}
\label{fig: cp_decomposition}
\end{figure}



\subsubsection{Tensor model for spatiotemporal travel pattern. } To preserve the innate hierarchical spatiotemporal multi-dimensional data, we further adopt a rank-1 CP tensor structure \citep{kolda2009tensor} for the function $P_k(\boldsymbol{w}_{u,i})$, where $\boldsymbol{w}=(o,d,t)$:  
\begin{equation}
P_k(\boldsymbol{w}_{u,i}) = P(w_{u,i}^O = o | z_u = k)P(w_{u,i}^D = d | z_u = k)P(w_{u,i}^T = t | z_u = k) = \phi_{k o}^O \phi_{k d}^D \phi_{k t}^T.     
\label{eq_pk_cp}
\end{equation}

From Eq. (\ref{eq_hier}), (\ref{eq_shorttxt}) and (\ref{eq_pk_cp}), the passenger likelihood is eventually represented as follows: 
\begin{equation}
\begin{split}
    P(\mathbf{d}_u) = \sum_{k} \theta_k \prod_{i=1}^{N_u} \phi_{k w_{u,i}^O}^O \phi_{k w_{u,i}^D}^D \phi_{k w_{u,i}^T}^T = [\![\boldsymbol{\theta},  \boldsymbol{\Phi}^O, \boldsymbol{\Phi}^D, \boldsymbol{\Phi}^T]\!]
\end{split}
\label{eq_cp}
\end{equation}

\textbf{CP Decomposition: }It is worth mentioning that Eq. (\ref{eq_cp}) reveals that the essence of the Tensor-DPMM model is the CP decomposition \citep{kolda2009tensor,ziyue2021tensor}, which decomposes the passenger likelihood $P(\mathbf{d}_u) \in \mathbb{R}^{V^O \times V^D \times V^T}$ is decomposed into the weight vector $\boldsymbol{\theta} \in \mathbb{R}^{K}$ and three mode matrices along each dimension of ODT, i.e., $\boldsymbol{\Phi}^O \in \mathbb{R}^{K \times V^O}$, $\boldsymbol{\Phi}^D \in \mathbb{R}^{K \times V^D}$, and $\boldsymbol{\Phi}^T \in \mathbb{R}^{K \times V^T}$, respectively.

When we introduce the Dirichlet Process to free $K$ from the parametric form, this is equalized by the work which conducts the CP decomposition with automatic determination of $K$ \citep{zhao2015bayesian,9126133}. \textcolor{black}{To the best of our knowledge, we are the first one who extends the model to a high-dimensional setting and applies it to intelligent transport systems.}


\subsection{Tensor Dirichlet Multinomial Mixture}
\label{sec: TensorDMM}

We will start with the basic version of the proposed model: a Tensor Dirichlet Multinomial Mixture (Tensor-DMM) model. 
Later we mainly emphasize the significant extensions based on this version. 

\begin{figure}[h]
\centering
\includegraphics[width=0.65\columnwidth]{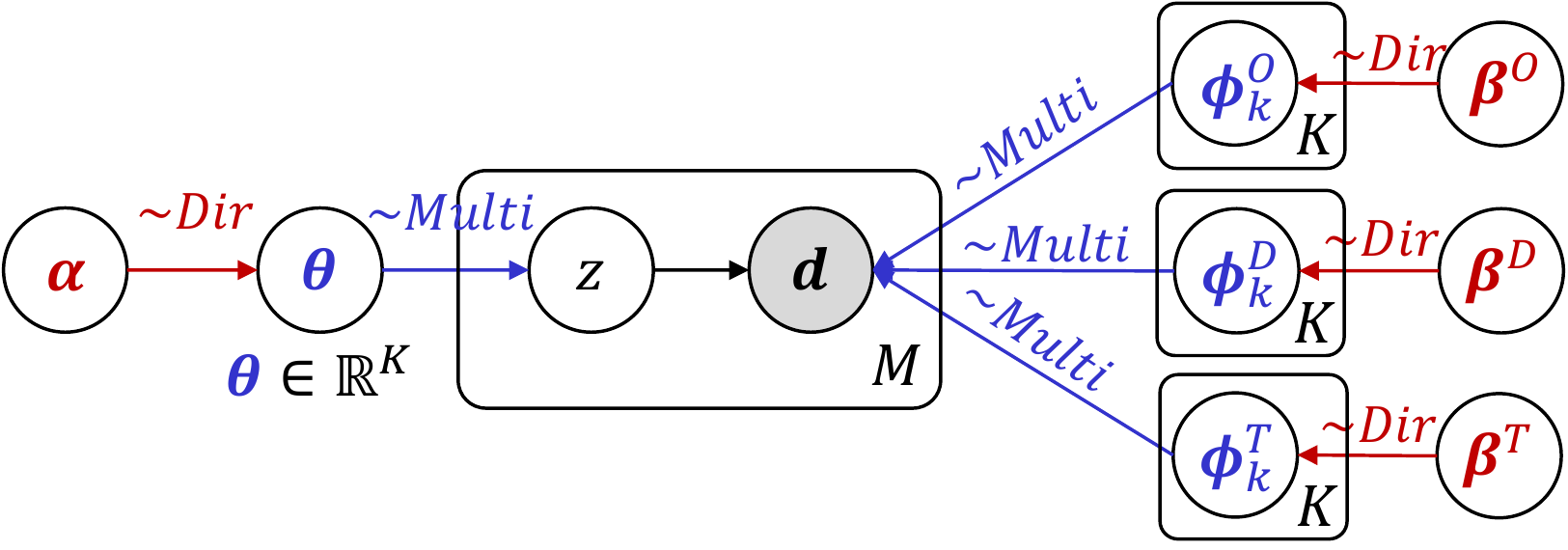}
\caption{Generative Process for Tensor-DMM Model}
\label{fig_gen_tensordmm}
\end{figure}

Tensor-DMM is designed as a generative model with multiple layers, which is suitable for the hierarchical trajectory data. We denote $P(z_u=k) = \theta_k$ and $\sum_{k} \theta_k=1$. 
The first layer generates a cluster $z_u$ via a multinomial distribution given passenger $u$ and the parameter $\boldsymbol{\theta}$ in Eq. (\ref{eq_gen_tensordmm2}). 
The second layer generates the trip elements $w^O, w^D, w^T$ via three multinomial distributions given the sampled cluster $z_u$ and the parameters $ \boldsymbol{\phi}_k^O, \boldsymbol{\phi}_k^D$, $\boldsymbol{\phi}_k^T$ in Eq. (\ref{eq_gen_tensordmm4}). 
The latent parameters $\boldsymbol{\theta}, \boldsymbol{\phi}_k^O, \boldsymbol{\phi}_k^D$, and $\boldsymbol{\phi}_k^T$ are drawn from Dirichlet distributions with prior parameters $\boldsymbol{\alpha}, \boldsymbol{\beta}^O, \boldsymbol{\beta}^D$, and $\boldsymbol{\beta}^T$, in Eq. (\ref{eq_gen_tensordmm1}) and Eq. (\ref{eq_gen_tensordmm3}), respectively. The whole generative process is shown in Figure \ref{fig_gen_tensordmm}.
\begin{align} 
  \label{eq_gen_tensordmm1}\tag{5.1}
  \boldsymbol{\theta} &\sim \text{Dir}(\boldsymbol{\alpha}) \\
  \label{eq_gen_tensordmm2}\tag{5.2}
  z_u &\sim \text{Multi}(\boldsymbol{\theta}), \forall u \in M \\
  \label{eq_gen_tensordmm3}\tag{5.3}
  \boldsymbol{\phi}_k^O \sim \text{Dir}(\boldsymbol{\beta}^O), \boldsymbol{\phi}_k^D &\sim \text{Dir}(\boldsymbol{\beta}^D), \boldsymbol{\phi}_k^T \sim \text{Dir}(\boldsymbol{\beta}^T), \forall k \in K \\
  \label{eq_gen_tensordmm4}\tag{5.4}
  w^O \sim \text{Multi}(\boldsymbol{\phi}_{z_u}^O), w^D &\sim \text{Multi}(\boldsymbol{\phi}_{z_u}^D), w^T \sim \text{Multi}(\boldsymbol{\phi}_{z_u}^T)
\end{align} 

\subsection{Tensor Dirichlet Process Multinomial Mixture}
\label{sec: TensorDPMM-G}
\label{sec: TensorDPMM}
\begin{figure}[h]
\centering
\includegraphics[width=0.65\columnwidth]{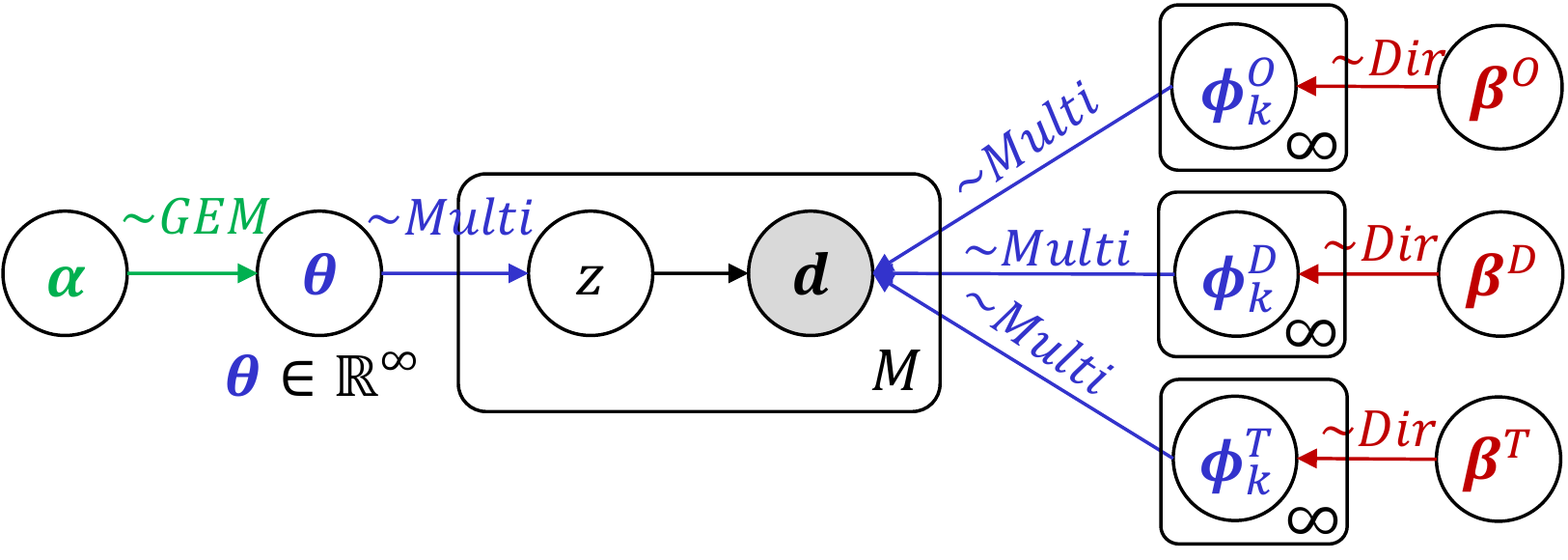}
\caption{Generative Process for Tensor-DPMM Model}
\label{fig_gen_tensordpmm}
\end{figure}

However, the Tensor-DMM model is formulated in a parametric form, with the number of clusters fixed as $K$. The proposed Tensor Dirichlet Process Multinomial Mixture (Tensor-DPMM) model instead relaxes the restriction and can be considered as an infinite extension of the Tensor-DMM. The Tensor-DPMM is summarized in Figure \ref{fig_gen_tensordpmm}.

The generative process for Tensor-DPMM is formulated as follows:

\begin{align} 
  \label{eq_gen_tensordpmm1}\tag{6.1}
  \boldsymbol{\theta} &\sim \text{GEM}(1, \boldsymbol{\alpha}) \\
  \label{eq_gen_tensordpmm2}\tag{6.2}
  z_u &\sim \text{Multi}(\boldsymbol{\theta}), \forall u \in M \\
  \label{eq_gen_tensordpmm3}\tag{6.3}
  \boldsymbol{\phi}_k^O \sim \text{Dir}(\boldsymbol{\beta}^O), \boldsymbol{\phi}_k^D &\sim \text{Dir}(\boldsymbol{\beta}^D), \boldsymbol{\phi}_k^T \sim \text{Dir}(\boldsymbol{\beta}^T), \forall k \in \infty \\
  \label{eq_gen_tensordpmm4}\tag{6.4}
  w^O \sim \text{Multi}(\boldsymbol{\phi}_{z_u}^O), w^D &\sim \text{Multi}(\boldsymbol{\phi}_{z_u}^D), w^T \sim \text{Multi}(\boldsymbol{\phi}_{z_u}^T)
\end{align} 

The Tensor-DPMM model replaces Dirichlet prior $\text{Dir}(\boldsymbol{\alpha})$ with a stick-breaking construction \citep{teh2007stick}, $\boldsymbol{\theta} \sim \text{GEM}(1, \boldsymbol{\alpha})$ in Eq. (\ref{eq_gen_tensordpmm1}). \textcolor{black}{The $\text{GEM}(\cdot)$ denotes the Stick-breaking construction for Dirichlet Process, and the letter stands for the names of the creators, Griffiths, Engen and McCloskey \citep{teh2010dirichlet}. The construction of $\boldsymbol{\theta}$ is illustrated via the stick-breaking analogy: starting with a stick with length $=1$, we break it for the first time and assign $\theta_1$ as the length of the sub-stick we just broke off. Similarly, we recursively break the rest portion to get $\theta_2, \theta_3$ and so on.}  As a result, the number of clusters is emerging with data dynamically. 

\section{Proposed Gibbs Sampling Algorithm}
\label{sec: algo_combined}
\textcolor{black}{As mentioned before, the generative processes of Tensor-DMM and Tensor-DPMM have the conjugate property since Dirichlet distribution is the conjugate prior to Multinomial distribution. Thus, Gibbs sampling will be adopted to learn the cluster. 
In Gibbs sampling, all the passengers will be initialized with an arbitrary cluster, and then in each iteration, the passenger will be "deselected" from the last iteration's cluster and then choose a new cluster with some given probabilities. The section \ref{sec: two_prob} will carefully derive the probabilities. And the section \ref{sec: algo} will give the final algorithm based on the formerly derived probabilities. }

\subsection{To Derive the Probability of Choosing A Cluster}
\label{sec: two_prob}
\textcolor{black}{We will first emphasize the difference between the cluster choosing of Tensor-DMM and Tensor-DPMM:} In the Tensor-DMM model, the Multinomial parameter to choose a cluster has a fixed dimension $\boldsymbol{\theta} \in \mathbb{R}^K$; However, in the Tensor-DPMM model, this parameter grows with the data $\boldsymbol{\theta} \in \mathbb{R}^\infty$, making the probability of a passenger choosing a cluster also dynamically evolve with data. \textcolor{black}{In short, Tensor-DMM could be regarded as a `simplified' version of Tensor-DPMM without a choice of creating a new cluster. Thus, this section will first derive the probabilities for Tensor-DPMM first in detail and then give the probability for Tensor-DMM directly.} 

\subsubsection{For Tensor-DPMM}
The core of the Dirichlet process, known as the Chinese Restaurant Process, is to define the following two probabilities: (1) choosing an existing cluster and (2) choosing a new cluster. These two probabilities will be further used to assign a cluster to each passenger during the proposed Gibbs Sampling. 

\paragraph{\textbf{(1) Choosing An Existing Cluster:}}
\label{sec: pz}
The probability of the passenger $\mathbf{d}_{u}$ choosing an existing cluster $z$ given the other passengers and their clusters is as follows, with details in the Appendix \ref{appdx: pz}:
\setcounter{equation}{6}
\begin{equation}\label{eq_prob_choosez}
\begin{split}
    & P(z_u = z | \mathbf{z}_{-u}, \mathcal{D}, \mathbf{\alpha}, \mathbf{\beta}^O, \mathbf{\beta}^D, \mathbf{\beta}^T)
    \\
    & \propto P(z_u = z | \mathbf{z}_{-u}, \mathcal{D}_{-u}, \mathbf{\alpha}, \beta^O, \beta^D, \beta^T) \times P(\mathbf{d}_{u} | z_u = z, \mathbf{z}_{-u}, \mathcal{D}_{-u}, \mathbf{\alpha}, \beta^O, \beta^D, \beta^T)\\
    &\propto P(z_u = z | \mathbf{z}_{-u}, \mathbf{\alpha}) \times P(\mathbf{d}_{u} | z_u = z, \mathcal{D}_{z, -u}, \beta^O, \beta^D, \beta^T),
\end{split}
\end{equation}
\noindent where $\mathcal{D}_{-u}$ ($\mathcal{D}_{z, -u}$) denotes the set including all the passengers (in the cluster $z$) except the $u$-th one.  
The first and second terms in Eq. (\ref{eq_prob_choosez}) are derived prudently.

The first term is solved as follows, with details in Appendix \ref{appdx: pz_term1}:
\begin{equation} \label{eq_choosez_term1}
    P(z_u = z | \mathbf{z}_{-u}, \mathbf{\alpha}) = \frac{m_{z,-u} + \frac{\alpha}{K}}{M-1+\alpha}.
\end{equation}
When $K \gg \alpha$, the prior probability in Eq. (\ref{eq_choosez_term1}) reduces to Eq. (\ref{eq_prel_choosek}).

The second term is solved as follows, with details in Appendix \ref{appdx: pz_term2}:
\begin{equation} \label{eq_choosez_term2}
\begin{split}
    P(\mathbf{d}_{u} | z_u = z, \mathcal{D}_{z, -u}, \beta^O, \beta^D, \beta^T)
    \propto \prod_{ODT} \left\{ \frac{\prod_{w \in \mathbf{d}_u} \prod_{j=1}^{N^{w}_u} (n_{z,-u}^{w} + \mathbf{\beta} +j - 1) }{\prod_{i=1}^{N_u}(n_{z,-u} + V \beta + i - 1)} \right\},
\end{split}
\end{equation}
where $N^{w}_u$ is the number of occurrences of trip $w$ in the $u$-th passenger regardless ODT; $n_{z,-u}^{w}, n_{z,-u}$ are the number of occurrences of trip $w$ in the $u$-th passenger and the total number of trips in cluster $z$ without considering the $u$-th passenger, respectively, and $n_{z,-u} = \sum_{w=1}^{V} n_{z,-u}^{w}$; $\prod_{ODT}\{\cdot\}$ means the product of the terms inside the bracket over all the dimensions of ODT.

Eq. (\ref{eq_choosez_term2}) shows that, in a high-dimensional case, the second term of Eq. (\ref{eq_prob_choosez}) is the decomposition result over all dimensions ODT.

Finally, the probability of the $u$-th passenger choosing the existing cluster $z$ given the other passengers and their clusters is:
\begin{equation}\label{eq_prob_choosez_final_1}
\begin{split}
    P_z &= P(z_u = z | \mathbf{z}_{-u}, \mathcal{D}, \mathbf{\alpha}, \mathbf{\beta}^O, \mathbf{\beta}^D, \mathbf{\beta}^T)  \propto \frac{m_{z,-u} + \frac{\alpha}{K}}{M-1+\alpha} \prod_{ODT} \left\{ \frac{\prod_{w \in \mathbf{d}_u} \prod_{j=1}^{N^{w}_u} (n_{z,-u}^{w} + \mathbf{\beta} +j - 1) }{\prod_{i=1}^{N_u}(n_{z,-u} + V \beta + i - 1)} \right\}.
\end{split}
\end{equation}
With the generalization of Tensor-DPMM that $K$ goes to infinity, we could get:

\begin{equation}\label{eq_prob_choosez_final}
\begin{split}
    P_z & \propto \frac{m_{z,-u}}{M-1+\alpha} \prod_{ODT} \left\{ \frac{\prod_{w \in \mathbf{d}_u} \prod_{j=1}^{N^{w}_u} (n_{z,-u}^{w} + \mathbf{\beta} +j - 1) }{\prod_{i=1}^{N_u}(n_{z,-u} + V \beta + i - 1)} \right\}.
\end{split}
\end{equation}

In Eq. (\ref{eq_prob_choosez_final}), the first term increases with a larger $m_{z,-u}$, which reflects the Rule-1 of ``choosing a cluster with more members''; The second term instead 
measures the ratio of the trip in the passenger $u$ over the trip in the cluster $z$, and it increases when cluster $z$ has more same trips with the passenger $u$, which reflects the Rule-2 of ``choosing a cluster with more commonalities'' \citep{yin2016model}. \textcolor{black}{It is also worth mentioning that the second term contains the information for all the ODT dimensions, contributing to another difference when we extend one-dimensional DPMM to Tensor-DPMM.}

\paragraph{\textbf{(2) Choosing A New Cluster:}} 
\label{sec: pK+1}
To choose a new cluster, the number of clusters increases to $K+1$. Similarly, the corresponding probability of the $u$-th passenger choosing a new cluster is as follows, with details in Appendix \ref{appdx: pnew}:
\begin{equation} \label{eq_prob_choose_new}
\begin{split}
    & P(z_u = K+1 | \mathbf{z}_{-u}, \mathcal{D}, \mathbf{\alpha}, \beta^O, \beta^D, \beta^T) \\
    & \propto P(z_u = K+1 | \mathbf{z}_{-u}, \mathbf{\alpha}) \times P(\mathbf{d}_{u} | z_u = K+1, \mathcal{D}_{z, -u}, \beta^O, \beta^D, \beta^T).
\end{split}
\end{equation}

The first term is derived as follows, with details in Appendix \ref{appdx: pnew_term1}:
\begin{equation} \label{eq_prob_choose_new_term1}
\begin{split}
    & P(z_u = K+1 | \mathbf{z}_{-u}, \mathbf{\alpha}) 
    = \frac{\alpha}{M-1+\alpha}.
\end{split}
\end{equation}

The second term is derived as follows, with details in Appendix \ref{appdx: pnew_term2}:
\begin{equation} \label{eq_prob_choose_new_term2}
\begin{split}
    P(\mathbf{d}_{u} | z_u = K+1, \mathcal{D}_{z, -u}, \beta^O, \beta^D, \beta^T)
    \propto \prod_{ODT} \left\{ \frac{\prod_{w \in \mathbf{d}_u} \prod_{j=1}^{N_u^{w}} (\beta + j -1)}{\prod_{i=1}^{N_u} (V \beta + i - 1)} \right\}.
\end{split}
\end{equation}

Thus, the final probability of passenger $u$ choosing a new cluster $K+1$ is:
\begin{equation} \label{eq_prob_choose_new_final}
\begin{split}
    P_{K+1} & = P(z_u = K+1 | \mathbf{z}_{-u}, \mathcal{D}, \mathbf{\alpha}, \beta^O, \beta^D, \beta^T) 
     \propto \frac{\alpha}{M-1+\alpha} \prod_{ODT} \left\{ \frac{\prod_{w \in \mathbf{d}_u} \prod_{j=1}^{N_u^{w}} (\beta + j -1)}{\prod_{i=1}^{N_u} (V \beta + i - 1)} \right\}.
\end{split}
\end{equation}

In Eq. (\ref{eq_prob_choose_new_final}), the first term still follows Rule-1 of ``choosing a cluster with more members'', where $\alpha$ is the pseudo number of members in the new cluster. Therefore, with a larger $\alpha$, the passenger tends to choose a new cluster; The second term still follows Rule-2 of ``choosing a cluster with more commonalities'', where $\beta$ is the pseudo frequencies of each trip in the new cluster.

Take the $10$-th passenger in Figure \ref{fig: CRP}. (a) as an example: the $10$-th passenger could either choose an existing cluster or a new cluster. In the illustration,  a ``round table'' means a cluster, a diamond means a passenger, and $\theta_z$ is the prior probability to choose cluster $z$, as shown in the Eq. (\ref{eq_gen_tensordpmm1}).

\subsubsection{For Tensor-DMM}

\textcolor{black}{The probabilities for choosing a clustering from a fixed pool of K clusters are quite similar to the process of `choosing an existing cluster' of Tensor-DPMM, i.e., Eq. (\ref{eq_prob_choosez_final_1}). The prior term, i.e., the first term will be adapted to the first term of Eq. (\ref{eq_dmm}). The second term will still be the tensor extension with the information from ODT. Thus, the probability of choosing a specific cluster from $K$ fixed clusters for Tensor-DMM is:}

\begin{equation}\label{eq_dmm_prob_choosek}
\begin{split}
    P_z \propto \frac{m_{z,-u} + \alpha}{M-1+K\alpha} \prod_{ODT} \left\{ \frac{\prod_{w \in \mathbf{d}_u} \prod_{j=1}^{N^{w}_u} (n_{z,-u}^{w} + \mathbf{\beta} +j - 1) }{\prod_{i=1}^{N_u}(n_{z,-u} + V \beta + i - 1)} \right\}.
\end{split}
\end{equation}

\subsection{The Final Algorithm Frameworks}
\label{sec: algo}

\textcolor{black}{After developing the probabilities of choosing clsuters, the Gibbs sampling algorithms are developed formally in this section. Similar to previous section, we will start from the algorithm for Tensor-DPMM first, and then introduce the algorithm for the `simplified' version, i.e., Tensor-DMM.}

\subsubsection{Tensor GSDPMM Algorithm with Disband and Relocate}
Different from the traditional collapsed Gibbs sampling for DPMM, known as GSDPMM for short \citep{griffiths2004finding,porteous2008fast,yin2014dirichlet,yin2016model}, this section proposes a tensor version along with the operations of ``disband'' and ``relocate'' to avoid small-size clusters. %

Considering the operational cost in practice, the size of clusters should not be too small since it is too expensive to provide a customized operation for a niche group. Therefore, we introduce a minimal cluster size requirement to control the cluster resolution, denoted as $r$. Taking the metaphor of a Chinese restaurant, we require each table to seat at least $r$ customers. 

The biggest differences between the proposed algorithm and the original GSDPMM lie in the following two aspects:

\begin{itemize}
    \item The original GSDPMM has been extended to a tensor high-dimensional version;
    \item A critical step of ``disband and relocate'' has been proposed to avoid the uncontrollable increase of cluster numbers. 
\end{itemize}

We will first introduce the Tensor Collapsed Gibbs Sampling algorithm, followed by a detailed explanation of the ``disband and relocate''.

As shown in Algorithm \ref{algo1}, the proposed Tensor Collapsed Gibbs Sampling with Disband and Relocated algorithm lies on its multi-dimensional Gibbs sampling phase (Phase 2.1 to Phase 2.3) and ``disband and relocate'' phase (Phase 3). The input of the proposed algorithm is the $M$ documents $\{\mathbf{d}_u\}_{u=1}^{M}$, along with the prior parameters. The prior parameters are chosen via grid search as shown in the experiments in Section  \ref{sec: parameter_tuning}.

The outputs include the number of clusters $K$ and the passenger cluster assignment $z_k$. The algorithm updates the parameters $m_z \in \mathbb{R}^K$ (number of passengers in cluster $z$), $n_z \in \mathbb{R}^K$ (number of trips in cluster $z$), $n_z^{w^O} \in \mathbb{R}^{KV^O}$ (number of occurrences of element $w^O$ in cluster $z$), $n_z^{w^D}$, and $n_z^{w^T}$. The algorithm runs with three phases: (1) Initialization, (2) Gibbs Sampling, and (3) Disband and Relocate.

\noindent \textbf{Phase 1 - Initialization}: We assign all passengers with one single cluster $K = K_0 = 1$, then the passengers will proceed into phase 2 for shuffling through $max\_iter$ iterations.

\noindent \textbf{Phase 2 - Gibbs Sampling}: In the Gibbs Sampling phase, three sub-phases are involved: ``Kick-out'', ``Choose a cluster'', and ``Merge''.
\begin{itemize}
    \item \textbf{Phase 2.1: Kick-out}: Firstly, we will kick the passenger $u$ out of its cluster $z$, i.e., $m_z = m_z-1$ and $n_z = n_z - N_u$ (passenger level). Simultaneously, the ODT elements in this passenger will also be removed from its original cluster:  $n_z^{w^O} = n_z^{w^O} + N_u^{w^O}$ (trip level). Opposite to ``selecting a cluster'', this step is also known as ``de-select''.
    \item \textbf{Phase 2.2: Choose A Cluster}: The kicked-out $u$-th passenger in the last step will re-choose the best fit cluster, i.e., either joining an existing cluster with the probability $\{P_z\}_{z=1}^{K}$ in Eq.(\ref{eq_prob_choosez_final}) or creating a new cluster with the probability $P_{K+1}$ in Eq.(\ref{eq_prob_choose_new_final}). This step reveals the importance of the two probabilities we derived in Section  \ref{sec: two_prob}. \textcolor{black}{If a new cluster is chosen with $P_{K+1}$, then the number of clusters will be updated as $K = K +1$. Such that, the cluster amount $K$ is learned along with the data.} 
    \item \textbf{Phase 2.3: Merge to the Selected Cluster}: After getting the new cluster assignment $z$ in Phase 2.2, now the $u$-th passenger will join the newly-selected $z$-th cluster: $m_z = m_z+1$ and $n_z = n_z + N_u$ (passenger level). Simultaneously, the ODT elements in this passenger will also be removed from its original cluster:  $n_z^{w^O} = n_z^{w^O} + N_u^{w^O}$ (trip level). 
\end{itemize}

\begin{algorithm}[t]
\SetAlgoVlined
\textbf{Input}: $\{\mathbf{d}_u\}_{u=1}^{M}, r, \alpha, {\beta}^{O}, {\beta}^{D}, {\beta}^{T}, max\_iter$\\
\textbf{Output}: $K, z_u, \forall u$\\
\textbf{Phase 1: Initialization}:\\
$K = K_0 (= 1), m_z, n_z, n_z^{w^O}, n_z^{w^D}, n_z^{w^T} =0$\\
\For{$u = 1$ to $M$}{
    Sample a cluster for $u$: $z_u \sim \text{Multi}(\frac{1}{K})$,
    $m_z = m_z + 1, n_z = n_z + N_u$\\
    \For{$i = 1$ to $N_u$}{
        $n_z^{w^O} = n_z^{w^O} + N_u^{w^O}$, same for D,T}
    }
\textbf{Phase 2: Gibbs Sampling:}\\
\For{$iter = 1$ to $max\_iter$}{
    \For{$u = 1$ to $M$}{
        \textbf{Phase 2.1: ``Kick-out'': }\\
        $m_z = m_z - 1, n_z = n_z - N_u$\\
        \For{$i = 1$ to $N_u$}{
            $n_z^{w^O} = n_z^{w^O} - N_u^{w^O}$, same for D,T
        }
        \If{$n_z ==0$}{
            $K=K-1$
        }
        \textbf{Phase 2.2: ``Choose A Cluster'': }\\
        $z_u \sim \{P_z\}_{z=1}^{K}, P_{K+1}$ in Eq. (\ref{eq_prob_choosez_final}) for an existing cluster and Eq. (\ref{eq_prob_choose_new_final}) for a new cluster\\
        \textbf{Phase 2.3: ``Merge to the Selected Cluster'': }\\
        \If{$z = K+1$}{
            $K = K+1$
        }
        $m_z = m_z + 1, n_z = n_z + N_u$\\
        \For{$i = 1$ to $N_u$}{
            $n_z^{w^O} = n_z^{w^O} + N_u^{w^O}$, same for D,T
        }
    }
}
\textbf{Phase 3: ``Disband Small Clusters and Relocate'': }\\
\For{$\forall z^* \in \{ z^* | m_{z^*} < r \}$}{
    \textbf{Phase 3.1: ``Disband'': }\\
    $K=K-1 $, delete $m_{z^*}, n_{z^*}, n_{z^*}^{w^O}, n_{z^*}^{w^D}, n_{z^*}^{w^T}$, record $u^* \in \{ u^* | z_{u^*} = z^*\}$ \\}
\For{$u^*=1$ to $M^*$}{
    \textbf{Phase 3.2: ``Relocate to remaining clusters'': }\\
    $z_u \sim \text{normalize}(\{P_z\}_{z=1}^{K})$ in Eq. (\ref{eq_prob_choosez_final})\\
    \textbf{Phase 3.3: ``Merge'':}  same as Phase 2.3\\
    }
\caption{Tensor Collapsed Gibbs Sampling with Disband and Relocate}
\label{algo1}
\end{algorithm}

\begin{figure}[t]
\centering
\includegraphics[width=0.75\columnwidth]{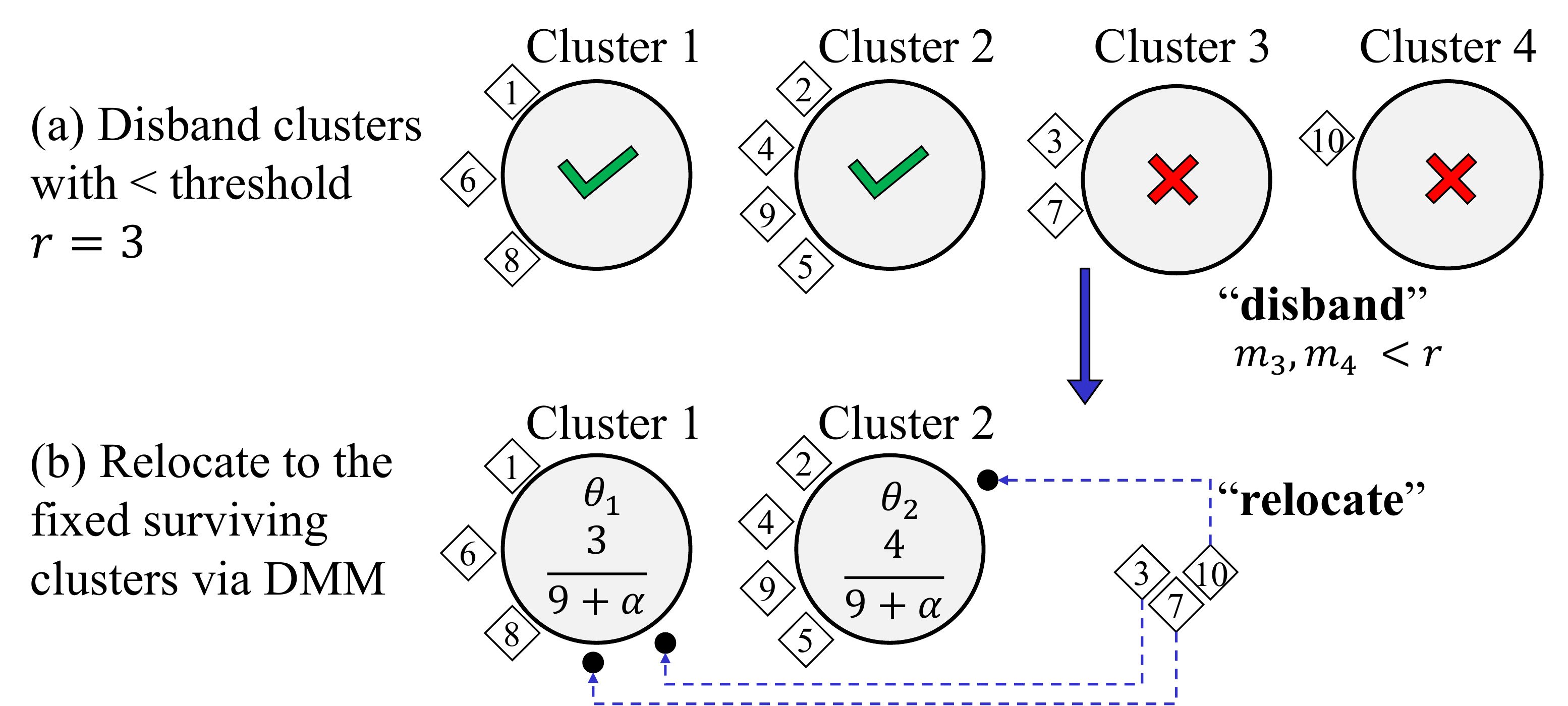}
\caption{(1) Disband clusters with $m_z < r$, e.g., $r=3$: cluster 3 and 4 are disbanded, $\{z^*\}=\{3,4\}$ and their members inside are $\{u^*\}=\{3,7,10\}$; (2) Relocate the members $\{u^*\}$ to the remaining clusters 1, 2 with probability $z_u \sim \text{normalize}(\{P_z\}_{z=1}^{K})$ (with a fixed $K=2$) in a DMM manner, without options to create a new cluster.}
\label{fig_disband_relocate}
\end{figure}

\noindent \textbf{Phase 3 - Disband and Relocate}:  
As mentioned before, after the regular Gibbs sampling phase, we innovatively propose a ``disband and relocate'' step after Phase 2 to check whether the formed clusters fulfill the minimal cluster size requirement, such that the total number of passenger clusters is at a controllable level. The phase of ``Disband and Relocate'' also has three sub-phases:
\begin{itemize}
    \item \textbf{Phase 3.1: Disband}: As illustrated in Figure \ref{fig_disband_relocate}. (a), we will disband the small clusters with $m_{z^*} < r$ in Phase 3.1, with the disbanded clusters denoted as $z^*$: for each $z^*$, $K=K-1$, and we delete its corresponding statistics, and we record all the member's insides $u^* = 1 \dots M^*$ ($M^*$ is the total population that is from the disbanded clusters), e.g., the member 3, 7, and 10 in the Figure \ref{fig_disband_relocate}. (a).
    \item \textbf{Phase 3.2: Relocate}: For the $M^*$ members from phase 3.1 that are to be related, we will relate them to the remaining surviving clusters. In this step, each member $u^*$ will face the choice to select a cluster again, like in Phase 2.2. However, in this relocating phase, to reduce the algorithm complexity and uncertainty, members are not given a chance to create a new cluster anymore. In other words, members could only choose a cluster from a fixed $K$ for the time being: $z_u \sim \text{normalize}(\{P_z\}_{z=1}^{K})$ in Eq. (\ref{eq_prob_choosez_final}). So this relocating step with a fixed $K$ (i.e., without a chance to create a new cluster anymore) is essentially a DMM model. As shown in Figure \ref{fig_disband_relocate}. (b), members 3, 7, and 10 are assigned to the remaining cluster 1 or 2.
    \item \textbf{Phase 3.3: Merge to the Selected Cluster}: After $\{u^*\}$s choosing their clusters, same as the Phase 2.3, the merge step is conducted in the passenger level and trip level: $m_z = m_z+1$, $n_z = n_z + N_u$ (passenger level) and $n_z^{w^O} = n_z^{w^O} + N_u^{w^O}$ (trip level). 
\end{itemize}

As mentioned before, a higher $\alpha$ means a passenger is more likely to create a new cluster. And the ``disband and relocate'' step is essentially competing again the $\alpha$ to control the number of clusters.

The computational complexity of each iteration is $\mathcal{O} \left( n_{dim}MNK \right)$, where $n_{dim}$ is the number of dimensions in the data, \textit{i.e.}, $n_{dim} = 3$ in our ODT data. This linear time complexity guarantees faster learning than two-step clustering methods with quadratic time.


\subsubsection{Tensor Algorithm GSDMM}

\textcolor{black}{In term of the Gibbs sampling algorithm for Tensor-DMM, it is following exactly the same Phase 1 and Phase 2, except the following two differences:}
\begin{itemize}
    \item In the Phase 1 of Initialization, instead of being initialized as $K=1$ in Tensor-DPMM, Tensor-DMM initializes $K$ as a large enough value, ideally larger than the ground truth.
    \item In the Phase 2.2 of choosing a cluster, no $P_{K+1}$ will be considered. 
    \item Correspondingly, in the Phase 2.3, there is no operation of $K=K+1$.
\end{itemize}
\textcolor{black}{Therefore, the number of clusters in Tensor-DMM will only start from a large value and be non-increasing over iterations. On the contrary, Tensor-DPMM starts with $K=1$, with $K$ growing with data. To avoid too much duplicated information, the Tensor-GS-DMM algorithm will not be shown here, instead in Algorithm. \ref{algo2} in the Appendix \ref{app: tensor-gs-dmm}.}

\section{Experiment}
\label{sec: experiment}
In this section, we will implement the proposed model into Hong Kong metro passenger trajectory data and compare our method with benchmarks. \textcolor{black}{ Section \ref{sec: data} will introduce the data first. Section \ref{sec: graph} will give a further extension of the proposed Tensor-DPMM model with spatial semantic graphs considered in the data preparation. We will first define the graphs and then deploy community detection algorithm to find the communities. Section \ref{sec: eval}, \ref{sec: parameter_tuning}, and \ref{sec: benchmark} will introduce the evaluation metrics, parameter tuning, and benchmark methods, respectively. The last two sections, i.e., Section \ref{sec: improvement} and \ref{sec: visual} will interpret and visualize the improved performance,  respectively.}

\subsection{Dataset} 
\label{sec: data}

The trajectory data are chosen from 50,000 randomly picked passengers during the period of 01-Jan-2017 to 31-Mar-2017. The information of entry station $w^O$, entry time (with a timestamp in an hour) $w^T$, and exit station $w^D$ are kept for analysis. The average length of each passenger is $\bar{N}  \approx 134$. The Hong Kong metro system had 98 stations in 2017 and operated in 24 hours. Thus the vocabulary size for origin, destination, and time is 98, 98, and 24. The data information is summarized in Table \ref{dataset_info}.

\begin{table}[h]
\centering
\caption{Data Information}
\begin{tabular}[t]{lcc}
\toprule
Data & Dimension and Description\\
\midrule

Passenger (Training) & $M = 50,000$ \\
Passenger (Validation, Test) & 1000 \\
Average Length & $\bar{N} \approx 134$ (trips) \\
Origin, Destination & $V^O, V^D = 98$ (stations) \\
Time (Entry) & $V^T = 24$ (hours) \\

\bottomrule
\end{tabular}
\label{dataset_info}
\end{table}%



\subsection{A Further Extension with Spatial Semantic Graph Considered}
\label{sec: graph}

The essence of traditional short text clustering models is to mechanically put documents that have the exact same words into the same cluster. However, in our spatiotemporal trajectory data, spatial semantic graphs are observed in origin, and destination dimensions \citep{geng2019spatiotemporal,li2020tensor}. For instance, in the origin dimension, although $o_1$ and $o_2$ are different stations, if they are graph neighbors, passengers containing $o_1$ and passengers containing $o_2$ can be in the same cluster. 
Existing methods usually incorporated graphs into topic models as Laplacian regularization \citep{mei2008topic,ziyue2021tensor}, but this will break the conjugate and introduce computational complexity. 

We aim to incorporate graphs without breaking the conjugate. To this end, we propose to use graph community detection \citep{kim2015community} to first identify the communities in graphs, denoted as $s \in S$, where $S$ is the total amount of detected communities, and then replace the original OD elements $w^O, w^D$ with the spatial communities. The rest of the model remains the same. 
\subsubsection{Graph Definition}
We formulate two spatial binary graphs to capture the geographical proximity and the functional similarity as follows.


\textbf{Geographical Proximity Graph:} The distance from station $i$ to $j$ is calculated. The proximity graph is formulated as an "$h$-hop" binary graph: if the distance between station $i$ and $j$ is less than $h$-hops, two stations are adjacent. We set $h=4$ since a survey stated that passengers are willing to travel freely if two stations are only 4 hops away (less than 10 minutes travel time) since it takes two minutes on average to travel between two stations  \citep{li2020tensor}.
\begin{equation} \label{eq_gadj}
  \{\mathbf{G}_{adj}\}_{i,j}=\begin{cases}
              1, &  \text{hop distance}_{i,j} \leq h\\
              0, &  \text{hop distance}_{i,j} > h\\
            \end{cases}
\end{equation}

\textbf{Functional Similarity Graph:} The functionality of a station can be characterized by the surrounding Point of Interest (POI) vectors, indicating the amount of the corresponding service (school, shopping malls, office buildings, etc.) nearby the station. The functional similarity thus can be defined as cosine similarity between POI vectors of the stations, and we assume stations $i$ and $j$ have similar functionality if the cosine similarity of their POIs is larger than a threshold $\gamma$.

\begin{equation} \label{eq_gpoi}
  \{\mathbf{G}_{poi}\}_{i,j}=\begin{cases}
              1, & \text{cos\_sim} \left(\textbf{POI}_{i},  {\textbf{POI}_{j}} \right) \geq \gamma\\
              0, &  \text{otherwise}\\
            \end{cases}
\end{equation}

The graph community detection not only merges two same semantic locations together but also reduces dimensions for OD. This technique is critical when dealing with a large transport network: it can reduce location dimensions $V^O$ and $V^D$ (\textit{e.g.}, $\approx 10^3$) to $S$ ($\approx 10^1$), which relieves heavy computational burden.  

\begin{figure}[t]
\centering
\includegraphics[width=0.85\columnwidth]{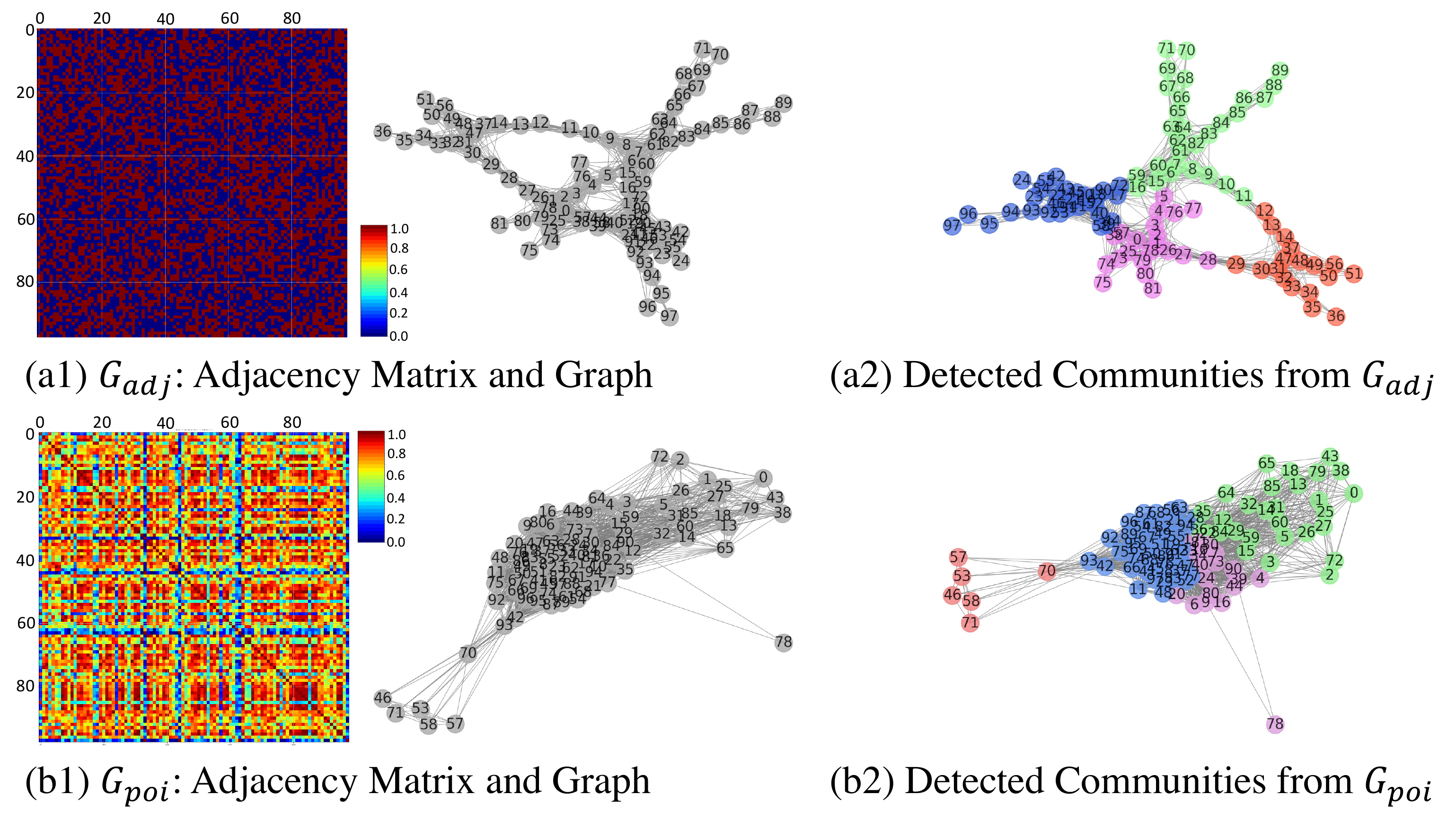}
\caption{(a1) The adjacency matrix and the graph visualization of $\mathbf{G}_{adj}$, with (a2) the detected four communities from $\mathbf{G}_{adj}$; (b1) The adjacency matrix and the graph visualization of $\mathbf{G}_{poi}$, with (a2) the detected four communities from $\mathbf{G}_{poi}.$}
\label{detected_community}
\end{figure}

\begin{figure}[t]
\centering
\includegraphics[width=0.65\columnwidth]{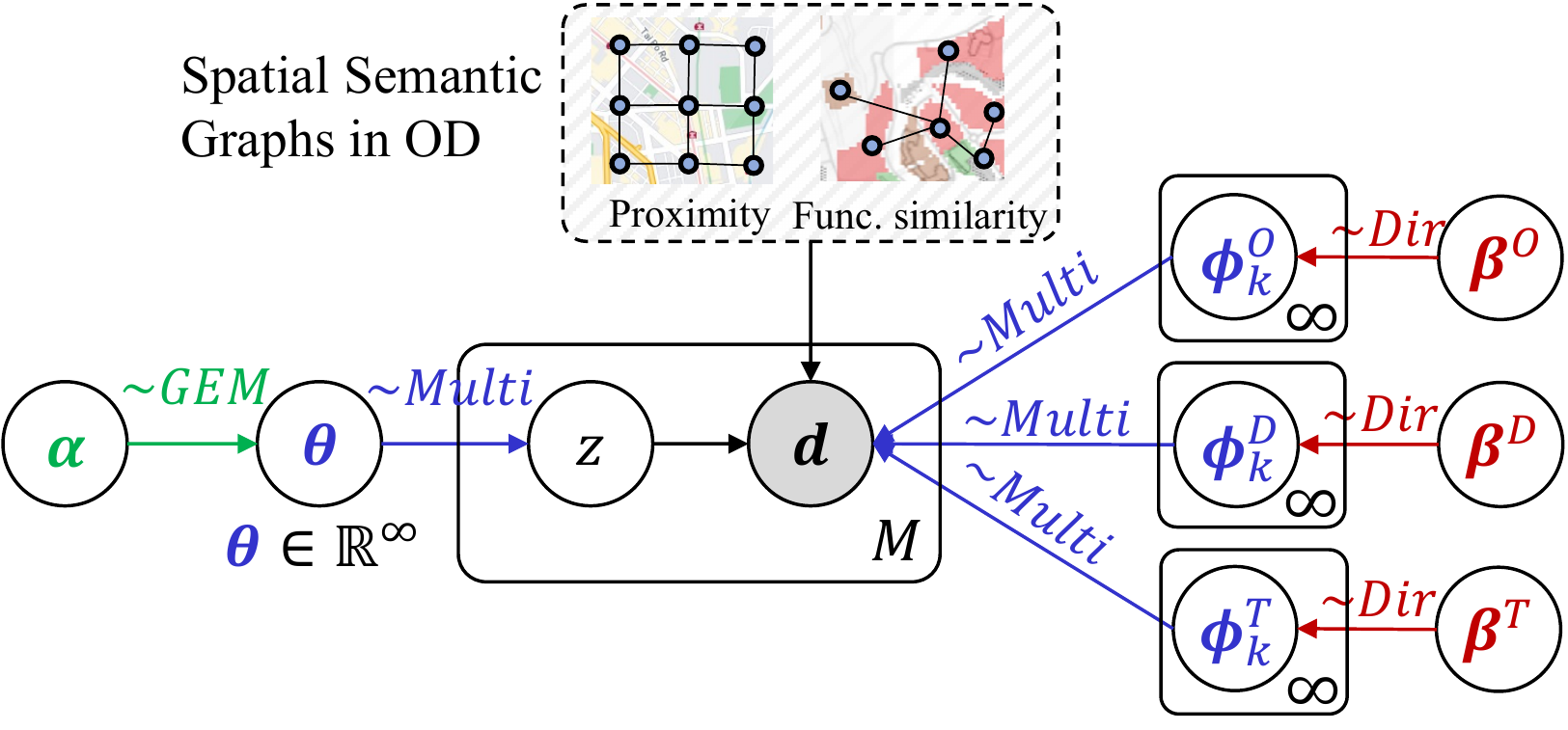}
\caption{The Tensor-DPMM-G Model}
\label{fig_gen_tensordpmm_g}
\end{figure}

\subsubsection{Graph Community Detection} 
Geographical proximity graph $\mathbf{G}_{adj}$ and functional similarity graph $\mathbf{G}_{poi}$ are observed in OD. In Eq. (\ref{eq_gadj}) and Eq. (\ref{eq_gpoi}), we set $h=4$ and $\gamma = 0.7$, such that the communities for each graph are detected with maximized modularity using the Leiden algorithm \citep{traag2019louvain}. Figure \ref{detected_community} shows four communities detected in both graphs. Then, each trip component $w^O, w^D$ will be replaced by their belonged communities, with community number as $4 \times 4$. After this pre-processing, the data will be fed into the proposed model. The new model variant is named as Tensor-DPMM-G, with the same generative process as Tensor-DPMM, and the only difference is whether the input data have OD information replaced with corresponding communities. We shown the Tensor-DPMM-G in Figure \ref{fig_gen_tensordpmm_g}.
%

\subsection{Evaluation Metrics} 
\label{sec: eval}

\begin{table}[t]
\centering
\caption{Resolution $r$ Sensitivity Analysis: Conducted with $\alpha, \beta^O, \beta^D=0.01, \beta^T = 0.042$}
\begin{threeparttable}
\setlength\tabcolsep{0pt} 
\begin{tabular*}{0.75\columnwidth}{@{\extracolsep{\fill}} lccccccc}
\toprule
$r$ & 15 & 25 & 35 & 45 & 55 & 65 & 75\\
\midrule
$K$ & 78 & 48 & 30 & 23 & 21 & 15 & 14\\
{\tiny RMSSTD} & 11.6 & 11.9 & 12.0 & \textbf{12.2} & 12.3 & 12.4 & 12.5
\\
RS & 1.57 & 1.56 & 1.54 & \textbf{1.52} & 1.51 & 1.49 & 1.47\\
CH & 0.71 & 0.72 & 0.74 & \textbf{0.88} & 0.73 & 0.69 & 0.55\\
\bottomrule
\end{tabular*}
\end{threeparttable}
\label{tab: r_search}
\end{table}

\begin{table}[t]
\centering
\caption{Prior $\alpha$ Sensitivity Analysis: Conducted with $r =45, \beta^O, \beta^D=0.01, \beta^T = 0.042$.}
\begin{threeparttable}
\setlength\tabcolsep{0pt} 
\begin{tabular*}{0.75\columnwidth}{@{\extracolsep{\fill}} lccccccc}
\toprule
$\alpha$ & 0.001 & 0.005 & 0.01 & 0.03 & 0.05 & 0.07 & 0.09\\
\midrule
$K$ & 26 & 25 & 23 & 25 & 21 & 26 & 25\\
{\tiny RMSSTD} & 12.3 & 12.3 & \textbf{12.2} & 12.5 & 12.4 & 12.5 & 12.4\\
RS & 1.50 & 1.50 & \textbf{1.52} & 1.49 & 1.50 & 1.49 & 1.50\\
CH & 0.70 & 0.73 & \textbf{0.88} & 0.68 & 0.74 & 0.68 & 0.76\\
\bottomrule
\end{tabular*}
\end{threeparttable}
\label{tab: alpha_search}
\end{table}

Since our clustering is conducted on real data without labels, traditional evaluation metrics, \textit{e.g.}, the normalized mutual information, are inapplicable due to the lack of ground truth. Thus, we use internal evaluation metrics in Eq. (\ref{evaluation}) to measure the compactness, separateness, and overall quality of clusters \citep{liu2010understanding}. Besides, we would also like to measure the run time.

\begin{itemize}
    \item \textbf{Mean run time} ($\bar{t}$): The average run time (in second $s$) per iteration.
    \item \textbf{Root-mean-square standard deviation} (RMSSTD): It measures the within-cluster compactness. Generally, a smaller RMSSTD value means a more compact cluster.
    \item \textbf{R-Squared} (RS): It measures cross-cluster separateness. Generally, a bigger RS means more separate different clusters.
    \item \textbf{\textit{Calinski}-\textit{Harabasz} index} (CH): It measures the two criteria above simultaneously by the formulation of $\frac{\text{separateness}}{\text{compactness}}$, which thus is recognized as the wholesome evaluation. We will use the CH index as the ultimate judge.
\end{itemize}
\begin{equation}
\begin{split}
    & \text{RMSSTD} = \left( \frac{\sum_{k} \sum_{u \in C_k} \|\mathbf{d}_u- \mathbf{c}_k \|^2}{V^O V^D V^T \sum_{k}(m_k-1)} \right)^{1/2},\\
    & \text{RS} = \frac{\sum_{u \in M} \| \mathbf{d}_u - \mathbf{g}\|^2 - \sum_k \sum_{u \in C_k} \|\mathbf{d}_u- \mathbf{c}_k \|^2}{\sum_{u \in M} \| \mathbf{d}_u - \mathbf{g}\|^2},\\
    & \text{CH} = \frac{\sum_{k} \|\mathbf{c}_k - \mathbf{g} \|^2 / (K-1)}{\sum_k \sum_{u \in C_k} \|\mathbf{d}_u- \mathbf{c}_k \|^2 / (M - K)},
\label{evaluation}
\end{split}
\end{equation}
where $C_k$ is the $k$-th cluster with its center  $\mathbf{c}_k$, and $\mathbf{g}$ is the global center of dataset. 

It is worth mentioning that the optimal value for RMSSTD and RS is not the smallest or biggest value; Instead, it is the first “elbow” in their monotonically increasing or decreasing curves \citep{hassani2017using}. Instead, the optimal value of CH is strictly maximal. Thus, the optimal clustering is chosen with the maximal CH.

\subsection{Best Parameter Searching and Sensitivity Analysis}
\label{sec: parameter_tuning}

\begin{figure}[t]
\centering
\includegraphics[width=0.6\columnwidth]{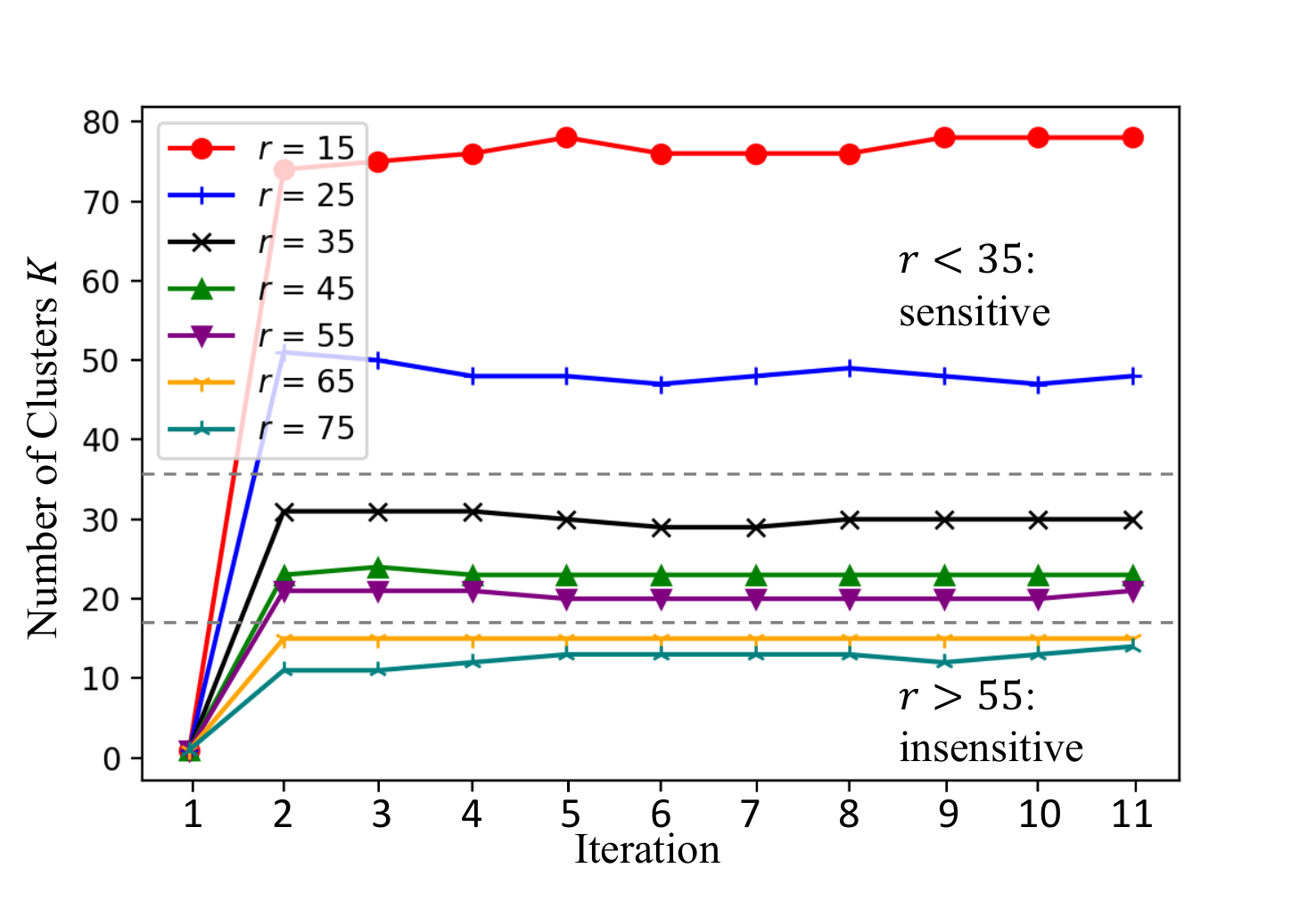}
\caption{Tensor-DPMM: Automatic learning process of the cluster number $K$ for Tensor-DPMM: Starting from $K=1$, the number of clusters is learned and growing along with data. The minimum cluster requirement $r$ also plays a role in $K$. A higher $r$ means a smaller $K$: when $r>55$, the $K$ is not affected by $r$ too much; whereas when $r<55$, $K$ is sensitive to $r$.}
\label{k_evolution}
\end{figure}

\begin{figure}[t]
\centering
\includegraphics[width=0.6\columnwidth]{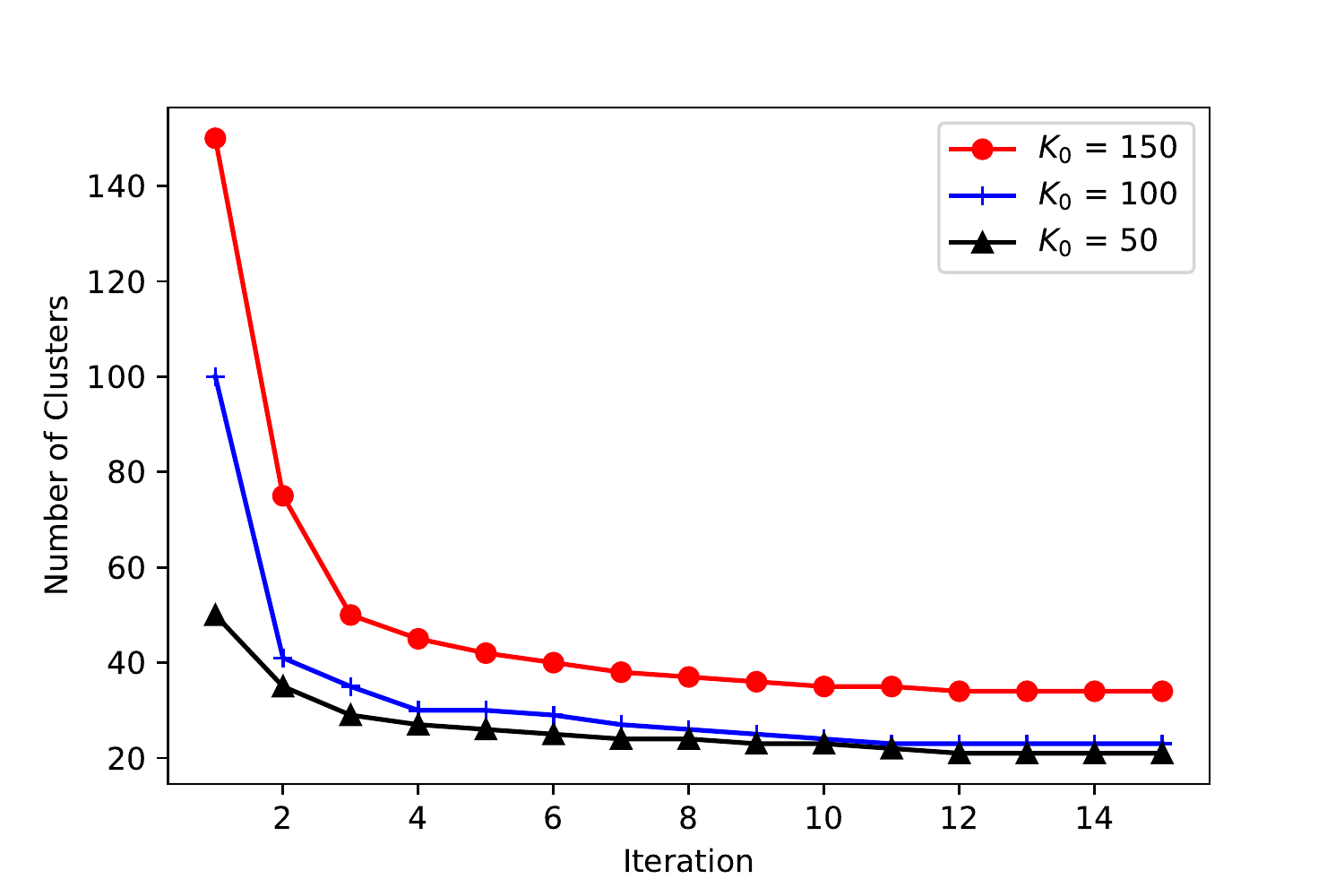}
\caption{\textcolor{black}{Tensor-DMM: Cluster Amount $K$ is non-increasing over iterations for Tensor-DMM. Different experiments with different start $K_0$ are conducted.}}
\label{k_evolution_dmm}
\end{figure}

The best set of parameters will be chosen via a grid search with the maximum CH index (or the first elbow of RMSSTD or RS value). In our case, $r=45, \alpha, \beta^O, \beta^D=0.01, \beta^T = 0.042$. Precisely, the minimum cluster size $r$ will directly affect the number of clusters: a larger $r$ will learn fewer clusters. Theoretically, $\alpha$ will also affect the number of clusters since a larger $\alpha$ means a passenger will be more likely to choose a new table. Therefore, we will mainly demonstrate how $r$ and $\alpha$ affect cluster learning, as shown in Table \ref{tab: r_search} and Table \ref{tab: alpha_search}.

As shown in Table \ref{tab: r_search}, RMSSTD monotonically increases and RS decreases with $r$ increasing (or $K$ decreasing), but the optimal $r=45$ is chosen with the maximum CH (same as the first elbow of RMSSTD and RS's monotonic trend) and $K=23$. 

\textbf{Tensor-DPMM: Automatic Learning of Cluster Amount:} The automatic and dynamic learning process of the number of clusters $K$ over iterations with different resolutions is shown in Figure \ref{k_evolution}. And $K$ is more sensitive with $r$ when $r<35$ and less sensitive when $r>55$. 

As shown in Table \ref{tab: alpha_search}, $K$ is not strictly increasing with $\alpha$. The main reason is that there are two rules mentioned before simultaneously affecting the choice of clusters. Besides, the ``disband and relocate'' step is also competing with a higher $\alpha$.

\textcolor{black}{\textbf{Tensor-DMM: A Large $K$ as Starting Point Needed:} For Tensor-DMM model, as mentioned before, the algorithm starts from a large enough $K$. Figure \ref{k_evolution_dmm} demonstrates the process.}

\subsection{Benchmark Methods} 
\label{sec: benchmark}
The following benchmark methods are selected.
\begin{itemize}
    \item \textbf{K-means}: It is one of the most common clustering methods, but it lacks the ability to cluster passengers whose trip records have multiple modes of ODT \citep{7891954}. Thus, we only use origin information to conduct this benchmark. 
    \item \textbf{Temporal Behaviour Clustering based on DBSCAN} (DBSCAN): Clustering based on passengers' temporal patterns, \textit{i.e.}, entry time, is widely used \citep{briand2017analyzing,he2020classification}. A mixture of unigrams \citep{mohamed2016clustering} model is used in this paper for comparison. Here we use DBSCAN \citep{ester1996density} to cluster the passengers' departing time.
    \item \textbf{DMM}: This short-text model \citep{yin2014dirichlet} can only handle one-dimensional words. To apply it, we either only utilize one dimension information (\textit{e.g.}, origin, denoted as DMM-O) or flatten three-dimensional $(w^O, w^D, w^T) = (o, d, t)$ trips into one-dimensional $w = odt$ (denoted as DMM-1d), which inevitably expands the vocabulary size exponentially to $V^O V^D V^T$.   
    \item \textbf{Graph-Regularized Tensor LDA} (GR-TensorLDA): It defines a generative process with trip-specific topics along each dimension of ODT. Graphs are added as Laplacian regularization \citep{ziyue2021tensor}. Passengers are clustered by distance-based methods using their probabilistic distributions of learned topics \citep{cheng2020probabilistic}. However, it is a two-step method with quadratic time, thus inefficient for clustering.
    \item \textbf{Tensor-DMM}: It is the basic version of the proposed model, which clusters passengers using their multi-dimensional trip records in a one-step manner.
    \item \textbf{Tensor-DPMM}: To validate the advantage of introducing spatial graphs, we also check the performance of the without-graph version.
\end{itemize}

We first compare all the methods by whether it is a tensor method to handle multi-dimensional data (\textit{Tensor}), cluster in a one-step manner (\textit{1-Step}), automatically determine $K$ (\textit{Auto-K}), and consider spatial semantic graphs (\textit{Graph}) in Table \ref{all_methods_comparison}. Only the proposed Tensor-DPMM-G model ticks all the boxes.

\begin{table}[t]
\centering
\caption{Internal Evaluation of Clustering Results from Different Methods.}\label{all_methods_comparison}
\begin{tabular}{c|cccc|cccc}
\toprule
Methods & \textit{Tensor} & \textit{1-Step} & \textit{Auto-$K$} & \textit{Graph} & $\bar{t} (s)$ & RMSSTD & RS & \textbf{CH} \\
\midrule
K-means &  $\times$ & $\checkmark$ & $\times$ & $\times$ & \textbf{5.000}  & 13.47 & 0.390 & 0.269 \\
DBSCAN & $\times$ & $\checkmark$ & $\times$ & $\times$ & 21.00 & 23.59 & \underline{1.491} & 0.215 \\
DMM-O & $\times$ & $\checkmark$ & $\times$ & $\times$ & 33.40 & 23.64 & 0.155 & 0.221 \\
DMM-1d & $\times$ & $\checkmark$ & $\times$ & $\times$ & 35.70 & \st{0.097} & 0.008 & 0.009 \\
\addlinespace
GR-TensorLDA & $\checkmark$ & $\times$ & $\times$ & $\checkmark$ & 8212 & 21.46 & 0.891 & 0.310 \\
\addlinespace
Tensor-DMM & $\checkmark$ & $\checkmark$ & $\times$ & $\times$ & 157.0 & 23.61 & 1.450 & 0.280 \\
Tensor-DPMM & $\checkmark$ & $\checkmark$ & $\checkmark$ & $\times$ & 1120 & 13.63 & 1.413 & \underline{0.676}\\
Tensor-DPMM-G & $\checkmark$ & $\checkmark$ & $\checkmark$ & $\checkmark$ & 73.2 
& \textbf{12.24} & \textbf{1.519} & \textbf{0.881} \\
\bottomrule
\end{tabular}
\end{table}

\subsection{Improved Performance} 
\label{sec: improvement}
\begin{figure}[t]
\centering
\includegraphics[width=\textwidth]{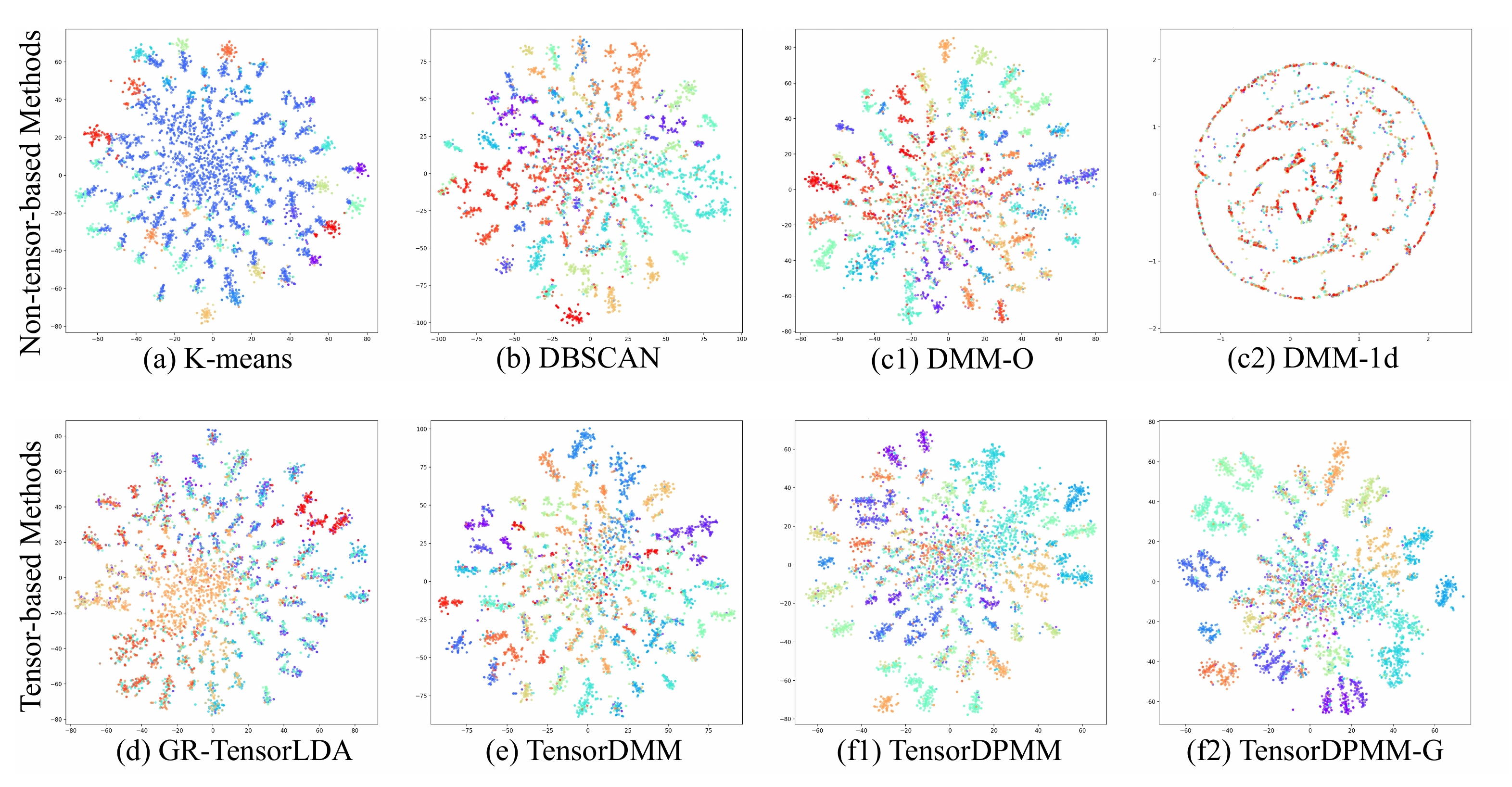}
\caption{Clustering Result Visualization via t-SNE \citep{van2008visualizing}, where different colors denote different clusters learned by each method} 
\label{cluster_visual}
\end{figure}

In Table \ref{all_methods_comparison}, the traditional one-mode methods  
(\textit{i.e.}, K-means, DBSCAN, and DMM) have lousy performance overall since the rich multi-mode spatiotemporal information is abandoned. Specifically, (1) K-means converges the fastest, yet learns a big blue cluster neighboring with other small clusters in Figure \ref{cluster_visual}. (a), which makes its relatively large RMSSTD yet small RS. This big cluster might be because some origins are popular for most of the passengers. (2) For DMM-1d in Figure \ref{cluster_visual}. (c2), when the ODT data is flattened as one-dimension, the vocabulary space expands to $V^O V^D V^T \approx 2\times10^5$, making the data extremely sparse and the worst clustering with the lowest CH. This flattening also changes the data distribution towards sparsity, as shown in Figure \ref{cluster_visual}. (d). This extreme sparsity makes each cluster quite compact within an extremely limited space, resulting in an abnormally small RMSSTD value. We do not consider this result for comparison. (3) DBSCAN based on the temporal feature only and DMM-O based on origin feature only also have unsatisfactory clustering results due to ignoring multi-mode spatiotemporal information, as mentioned before.

Tensor methods achieve overall better performance than non-Tensor methods due to well-preserved ODT information. However, GR-TensorLDA 
is sub-optimal with a lower CH value compared to ours since it is designed to learn local clusters in each ODT dimension, not a global cluster for passengers. Besides, since it is to learn the hidden cluster for each trip and its clustering is in a two-step manner, its run time is the highest.

The proposed Tensor-DPMM-G 
achieves the highest 
CH value, which is four times higher than the benchmarks, proving that the proposed model can learn clusters with better qualities. Precisely, (1) the Tensor-DPMM-G learns clusters with more within-cluster compactness (second lowest RMSSTD), more cross-cluster separateness (highest RS), and the best overall quality (highest CH). (2) Furthermore, compared with our Tensor-DPMM without graphs, incorporating graphs further improves the clusters. (3) In terms of run time, Tensor-DMM runs 50 times faster than GR-TensorLDA given the simplified short-text assumption and one-step clustering; Tensor-DPMM takes 10 times longer than Tensor-DMM since the stochastic Dirichlet process of ``choosing a cluster'' adds more uncertainty in the learning process, yet still 7 times faster than GR-TensorLDA; Tensor-DPMM-G speeds up 15 times more than Tensor-DPMM since graph community detection reduces the dimension of OD. \textcolor{black}{It is worth mentioning that the traditional single-mode methods still have a shorter run time than our Tensor-DPMM-G. This is because our high-dimensional methods' complexity is proportional to the dimensionality. Our method considers three dimensions. Thus, for a fair comparison, our run time per dimension is around 24.4s for each iteration, which is quite competitive among the single-mode methods.}

\subsection{Visualization}
\label{sec: visual}
\textcolor{black}{We also visualize the learned clusters via t-distributed Stochastic Neighbor Embedding (t-SNE) \citep{van2008visualizing}. 
The passenger trip data are input into t-NSE for embedding learning so that each passenger's travel data can be plotted as an instance in Figure  \ref{cluster_visual}. So far, we have used the origin dimension as input for demonstration purposes. The learned cluster assignments are another input to color different passengers, with a different color corresponding to a different cluster. The other dimensions have similar results.} 

\textcolor{black}{As shown in Figure \ref{cluster_visual}. (a), K-means learns a big blue cluster, as mentioned before. In Figure \ref{cluster_visual}. (b) and (c1), DBSCAN and DMM-O have similar mixed cluster distributions, representing the temporal feature and the spatial (origin) feature, respectively. This also proves the necessity to well preserve the spatiotemporal multi-mode structure. In Figure \ref{cluster_visual}. (c2), DMM-1d's visualization experiences the longest processing time for t-NSE embedding due to the exponentially increased vocabulary size by folding three dimensions into one. Its visualization also suffers from extreme sparsity. In Figure \ref{cluster_visual}. (d), GR-TensorLDA could roughly separate the orange cluster and the blue/purple clusters, which is a good improvement compared to the single-mode methods; However, within each cluster, there are instances of other colors (cluster). This may be because the GR-TensorLDA is designed to learn the local clustering structure of each mode of ODT instead of learning a global clustering structure for the passenger level. In Figure \ref{cluster_visual}. (e), our Tensor-DMM, specialized to learn a global clustering for passengers, clearly achieves more pure clustering than GR-TensorLDA. However, as mentioned before, Tensor-DMM lacks desired flexibility since it needs a specified and fixed $K$.  In Figure \ref{cluster_visual}. (f1) and (f2), our Tensor-DPMM and its variant with graph, Tensor-DPMM-G, could so far achieve the most clear-cut clustering results. Comparing these two versions, we could also find that after considering spatial graphs, the clusters in the same color (e.g., purple, dark blue, light green, etc.) are learned closer.}

In summary, Tensor-DPMM-G strikes a balance of offering the best clustering with relatively fast enough computation.

\section{Conclusion and Future Work}
\label{sec: conclude}
This paper studied a passenger clustering problem based on hierarchical, multi-dimensional, and spatiotemporal trip data. We proposed to use the tensor to preserve the spatiotemporal data, and we proposed \textcolor{black}{ two model, namely, Tensor Dirichlet Multinomial Mixture (Tensor-DMM) model and Tensor Dirichlet Process Multinomial Mixture (Tensor-DPMM) model to cluster the passenger in a one-step manner. The Tensor-DPMM model specially could automatically learn the cluster number. We further extend the Tensor-DPMM model with two graphs considered, namely, the geographical and the functional graphs.} In addition, we offered a tensor Gibbs sampling algorithm with a minimum cluster size requirement via an innovative ``disband and relocate'' step. In the case study, we conducted prudent experiments and demonstrated the improved qualities of the learned clusters, with four times higher CH value and fast enough computation. 

For future research, we aim to embed graphs directly into the model's generative process as a unified model. Besides, we also plan to add trip duration as another dimension.

\section*{Acknowledgement}
This work is partially funded with the RGC GRF 16201718 and 16216119. The authors appreciate the help from the Hong Kong MTR Co. research, marketing, and customer service teams. The authors would also like to thank all the reviewers and editors for their precious suggestions.

\newpage
\appendix
\section{Appendix: Choosing A Cluster}
\label{appx: choose-cluster}
In this section, we will derive the two following probabilities in detail based on Dirichlet Process: (1) choosing an existing cluster, denoted as $P_z$, and (2) choosing a new cluster, denoted as $P_{K+1}$. To avoid confusion, the equation numbering in this appendix is kept consistent with the equations that appear in the main paper.

\subsection{Choosing An Existing Cluster $P_z$:} \label{appdx: pz}

The probability of the passenger $\mathbf{d}_{u}$ choosing an existing cluster $z$ given the other passengers and their clusters is:

\begin{equation}\label{pz}\tag{7}
\begin{split}
    & P_z = P(z_u = z | \mathbf{z}_{-u}, \mathcal{D}, \mathbf{\alpha}, \beta^O, \beta^D, \beta^T) \\
    & \propto P(z_u = z | \mathbf{z}_{-u}, \mathcal{D}_{-u}, \mathbf{\alpha}, \beta^O, \beta^D, \beta^T) \times P(\mathbf{d}_{u} | z_u = z, \mathbf{z}_{-u}, \mathcal{D}_{-u}, \mathbf{\alpha}, \beta^O, \beta^D, \beta^T)\\
    &\propto P(z_u = z | \mathbf{z}_{-u}, \mathbf{\alpha}) \times P(\mathbf{d}_{u} | z_u = z, \mathcal{D}_{z, -u}, \beta^O, \beta^D, \beta^T)
\end{split}
\end{equation}
\noindent where $\mathcal{D}_{-u}$ ($\mathcal{D}_{z, -u}$) is all the passengers (in the cluter $z$) except the $u$-th one.  

We will derive the first term $P(z_u = z | \mathbf{z}_{-u}, \mathbf{\alpha}) $ and the second term $P(\mathbf{d}_{u} | z_u = z, \mathcal{D}_{z, -u},  \beta^O, \beta^D, \beta^T)$ with Appx. \ref{appdx: pz_term1} and Appx. \ref{appdx: pz_term2} respectively.

\subsubsection{The First Term of $P_z$} \label{appdx: pz_term1}

Borrowing the results from \citep{yin2016model}, the first term could be written as follows:

\begin{equation}\label{pz_term1}\tag{b1.1}
    P(z_u = z | \mathbf{z}_{-u}, \mathbf{\alpha}) = \int P(\mathbf{z}_{u}=z | \boldsymbol{\theta}) P(\boldsymbol{\theta} | \mathbf{z}_{-u}, \alpha)  d\boldsymbol{\theta} 
\end{equation}
where $P(\mathbf{z}_{u}=z | \boldsymbol{\theta})$ can be easily given as:

\begin{equation}\label{pz_term11}\tag{b1.1.1}
    P(\mathbf{z}_{u}=z | \boldsymbol{\theta}) = \text{Multi}(\mathbf{z}_{u}=z | \boldsymbol{\theta})= \theta_z
\end{equation}

However, $P(\boldsymbol{\theta} | \mathbf{z}_{-u}, \alpha)$ needs more derivations as follows \citep{yin2016model}:

\begin{equation}\label{pz_term12}\tag{b1.1.2}
\begin{split}
    P(\boldsymbol{\theta} | \mathbf{z}_{-u}, \alpha) 
    & = \frac{P(\boldsymbol{\theta} | \alpha) P(\mathbf{z}_{-u}|\boldsymbol{\theta})}{\int P(\boldsymbol{\theta} | \alpha) P(\mathbf{z}_{-u}|\boldsymbol{\theta}) d\boldsymbol{\theta} } = \frac{ \frac{1}{\Delta(\alpha)} \prod_{k=1}^K \boldsymbol{\theta}_k^{\frac{\alpha}{K} - 1} \prod_{k=1}^K \boldsymbol{\theta}_k^{m_{k,-u}} }{ \int \frac{1}{\Delta(\alpha)} \prod_{k=1}^K \boldsymbol{\theta}_k^{\frac{\alpha}{K} - 1} \prod_{k=1}^K \boldsymbol{\theta}_k^{m_{k,-u}} d\boldsymbol{\theta} } \\
    & = \frac{1}{\Delta(\mathbf{m}_{-d} + \frac{\alpha}{K})} \prod_{k=1}^K \boldsymbol{\theta}_k^{ m_{k,-u} + \frac{\alpha}{K} - 1} = \text{Dir}( \boldsymbol{\theta} | \mathbf{m}_{-d} +  \frac{\alpha}{K})
\end{split}
\end{equation}

We substitute Eq. (\ref{pz_term11}) and  Eq. (\ref{pz_term12}) into Eq. (\ref{pz_term1}):

\begin{equation}\label{pz_term1_final}\tag{8}
\begin{split}
    P(z_u = z | \mathbf{z}_{-u}, \mathbf{\alpha})
    & = \int \text{Multi}(\mathbf{z}_{u}=z | \boldsymbol{\theta}) \times \text{Dir}( \boldsymbol{\theta} | \mathbf{m}_{-d} +  \frac{\alpha}{K}) d\boldsymbol{\theta} \\
    & = \int \theta_z \frac{1}{\Delta(\mathbf{m}_{-d} + \frac{\alpha}{K})} \prod_{k=1, k \neq z}^K \boldsymbol{\theta}_k^{m_{k,-u} + \frac{\alpha}{K} -1} d\boldsymbol{\theta} \\
    & = \frac{ \Delta(\mathbf{m} + \frac{\alpha}{K}) }{ \Delta(\mathbf{m}_{-d} + \frac{\alpha}{K}) } = \frac{ \prod_{k=1}^K \Gamma(m_k+\frac{\alpha}{K}) }{ \Gamma(\prod_{k=1}^K (m_k+\frac{\alpha}{K})) } \frac{ \Gamma( \prod_{k=1}^K (m_{k,-u}+\frac{\alpha}{K}) ) }{ \prod_{k=1}^K \Gamma( m_{k,-u}+\frac{\alpha}{K} ) } \\
    & = \frac{ \Gamma( m_{z,-u}+\frac{\alpha}{K} + 1) }{\Gamma( m_{z,-u}+\frac{\alpha}{K} )} \frac{ \Gamma(M - 1 + \alpha) }{\Gamma(M + \alpha )} \\
    & = \frac{m_{z,-u} + \frac{\alpha}{K}}{M-1+\alpha}
\end{split}
\end{equation}
where $\Delta (\boldsymbol{\alpha}) = \frac{ \prod_{k=1}^K \Gamma(\alpha) }{ \Gamma(\prod_{k=1}^K \alpha) } $, $\Gamma(\cdot)$ is the gamma function, and $\Gamma(x+1) = x \Gamma(x)$.

\subsubsection{The Second Term of $P_z$} \label{appdx: pz_term2}

The second term in Eq. (\ref{pz}) is expanded as follows:

\begin{equation}\label{pz_term2}\tag{b1.2}
\begin{split}
    & P(\mathbf{d}_{u} | z_u = z, \mathcal{D}_{z, -u}, \mathbf{\beta}^O, \mathbf{\beta}^D, \mathbf{\beta}^T) \\
    & = \int P( \mathbf{d}_{u}, \phi_z^O, \phi_z^D, \phi_z^T | z_u = z, \mathcal{D}_{z, -u}, \mathbf{\beta}^O, \mathbf{\beta}^D, \mathbf{\beta}^T ) d\phi_z^O d\phi_z^D d\phi_z^T \\
    & = \int P( \phi_z^O | \mathcal{D}_{z, -u}, \mathbf{\beta}^O )  P( \phi_z^D | \mathcal{D}_{z, -u}, \mathbf{\beta}^D ) P( \phi_z^T | \mathcal{D}_{z, -u}, \mathbf{\beta}^T ) \times P( \mathbf{d}_{u} | \phi_z^O, \phi_z^D, \phi_z^T, z_u = z ) d\phi_z^O d\phi_z^D d\phi_z^T
\end{split}
\end{equation}

Borrowing the result from \citep{yin2016model}, $P( \phi_z^O | \mathcal{D}_{z, -u}, \mathbf{\beta}^O )$ can be given as follows:

\begin{equation}\label{pz_term21}\tag{b1.2.1}
\begin{split}
    P( \phi_z^O | \mathcal{D}_{z, -u}, \mathbf{\beta}^O )
    & = \frac{ P( \phi_z^O | \beta^O ) P( \mathcal{D}_{z, -u}| \phi_z^O ) }{\int P( \phi_z^O | \beta^O ) P( \mathcal{D}_{z, -u}| \phi_z^O ) {\,\,} d\phi_z^O}  = \frac{ \frac{1}{\Delta(\beta^O)} \prod_{w^O} \phi_{z,w^O}^{\beta^O - 1} \prod_{w^O} \phi_{z,w^O}^{n_{z,-u}^{w^O}} }{ \int \frac{1}{\Delta(\beta^O)} \prod_{w^O} \phi_{z,w^O}^{\beta^O - 1} \prod_{w^O} \phi_{z,w^O}^{n_{z,-u}^{w^O}}  {\,\,} d\phi_z^O}\\
    & = \frac{1}{\Delta(\mathbf{n}_{z,-u} + \beta^O)} \prod_{w^O} \phi_{z,w^O}^{n_{z,-u}^{w^O} + \beta^O - 1} = \text{Dir}( \phi_z^O | \mathbf{n}_{z,-u} + \mathbf{\beta}^O)
\end{split}
\end{equation}

We have the same results for dimension D and T:

\begin{align*}
    P( \phi_z^D | \mathcal{D}_{z, -u}, \mathbf{\beta}^D ) = \text{Dir}( \phi_z^D | \mathbf{n}_{z,-u} + \mathbf{\beta}^D) \label{pz_term22}\tag{b1.2.2}\\
    P( \phi_z^T | \mathcal{D}_{z, -u}, \mathbf{\beta}^T ) = \text{Dir}( \phi_z^T | \mathbf{n}_{z,-u} + \mathbf{\beta}^T) \label{pz_term23}\tag{b1.2.3}\\
\end{align*}

$ P( \mathbf{d}_{u} | \phi_z^O, \phi_z^D, \phi_z^T, z_u = z )$ will be derived as follows:

\begin{equation}\label{pz_term24}\tag{b1.2.4}
\begin{split}
    P( \mathbf{d}_{u} | \phi_z^O, \phi_z^D, \phi_z^T, z_u = z )
    & = \prod_{i \in N_u} \text{Multi}(w_i^O | \phi_z^O) \text{ Multi}(w_i^D | \phi_z^D) \text{ Multi}(w_i^T | \phi_z^T) \\
    & = \prod_{i \in N_u} \phi_{z,w_i^O}^O \phi_{z,w_i^D}^D \phi_{z,w_i^T}^T  = \prod_{w^O} \phi_{z,w^O}^{n_u^{w^O}} \prod_{w^D} \phi_{z,w^D}^{n_u^{w^D}} \prod_{w^T} \phi_{z,w^T}^{n_u^{w^T}}
\end{split}
\end{equation}

We substitute all the terms from Eq. (\ref{pz_term21}), (\ref{pz_term22}), (\ref{pz_term23}), and (\ref{pz_term24}) to Eq. (\ref{pz_term2}) such that we could obtain the following result:

\begin{equation}\label{pz_term2_final}\tag{9}
\begin{split}
    & P(\mathbf{d}_{u} | z_u = z, \mathcal{D}_{z, -u}, \mathbf{\beta}^O, \mathbf{\beta}^D, \mathbf{\beta}^T) \\
    = & \int [ \text{Dir}( \phi_z^O | \mathbf{n}_{z,-u} + \mathbf{\beta}^O) \text{Dir}( \phi_z^D | \mathbf{n}_{z,-u} + \mathbf{\beta}^D) \text{Dir}( \phi_z^T | \mathbf{n}_{z,-u} + \mathbf{\beta}^T) \times \prod_{w^O} \phi_{z,w^O}^{n_u^{w^O}} \prod_{w^D} \phi_{z,w^D}^{n_u^{w^D}} \prod_{w^T} \phi_{z,w^T}^{n_u^{w^T}} ] {\,\,}d\phi_z^O d\phi_z^D d\phi_z^T \\
    = & \frac{ \Delta( \mathbf{n}_z + \beta^O ) }{ \Delta( \mathbf{n}_{z,-u} + \beta^O ) } \frac{ \Delta( \mathbf{n}_z + \beta^D ) }{ \Delta( \mathbf{n}_{z,-u} + \beta^D ) }  \frac{ \Delta( \mathbf{n}_z + \beta^T ) }{ \Delta( \mathbf{n}_{z,-u} + \beta^T ) } \\
    = & \frac{ \prod_{w^O} \Gamma(n_z^{w^O} + \beta^O) }{ \Gamma( \sum_{w^O} (n_z^{w^O} + \beta^O)) } \frac{ \Gamma( \sum_{w^O} (n_{z,-u}^{w^O} + \beta^O)) }{ \prod_{w^O} \Gamma(n_{z,-u}^{w^O} + \beta^O) }  \times \frac{ \prod_{w^D} \Gamma(n_z^{w^D} + \beta^D) }{ \Gamma( \sum_{w^D} (n_z^{w^D} + \beta^D)) } \frac{ \Gamma( \sum_{w^D} (n_{z,-u}^{w^D} + \beta^D)) }{ \prod_{w^D} \Gamma(n_{z,-u}^{w^D} + \beta^D) } \\
    & {\,\,}\times \frac{ \prod_{w^T} \Gamma(n_z^{w^T} + \beta^T) }{ \Gamma( \sum_{w^T} (n_z^{w^T} + \beta^T)) } \frac{ \Gamma( \sum_{w^T} (n_{z,-u}^{w^T} + \beta^T)) }{ \prod_{w^T} \Gamma(n_{z,-u}^{w^T} + \beta^T) } \\
    = & \frac{\prod_{w^O } \prod_{j=1}^{N^{w^O}_d} (n_{z,-d}^{w^O} + \mathbf{\beta}^O +j - 1) }{\prod_{i=1}^{N_u}(n_{z,-u} + V^O \beta^O + i - 1)}  \times \frac{\prod_{w^D } \prod_{j=1}^{N^{w^D}_d} (n_{z,-d}^{w^O} + \mathbf{\beta}^D +j - 1) }{\prod_{i=1}^{N_u}(n_{z,-u} + V^D \beta^D + i - 1)}   \times \frac{\prod_{w^T } \prod_{j=1}^{N^{w^T}_d} (n_{z,-d}^{w^T} + \mathbf{\beta}^T +j - 1) }{\prod_{i=1}^{N_u}(n_{z,-u} + V^T \beta^T + i - 1)} \\
    = & \prod_{ODT} \left\{ \frac{\prod_{w \in \mathbf{d}_u} \prod_{j=1}^{N^{w}_u} (n_{z,-u}^{w} + \mathbf{\beta} +j - 1) }{\prod_{i=1}^{N_u}(n_{z,-u} + V \beta + i - 1)} \right\}
\end{split}
\end{equation}

Finally, we substitute Eq. (\ref{pz_term1_final}) and (\ref{pz_term2_final}) into Eq. (\ref{pz}), then the probability of passenger choosing the existing cluster $z$ given the other passengers and their clusters is eventually obtained as:
\begin{equation}\label{pz}\tag{10}
\begin{split}
    P_z & = P(z_u = z | \mathbf{z}_{-u}, \mathcal{D}, \mathbf{\alpha}, \mathbf{\beta}^O, \mathbf{\beta}^D, \mathbf{\beta}^T)  \propto \frac{m_{z,-u} + \frac{\alpha}{K}}{M-1+\alpha} \prod_{ODT} \left\{ \frac{\prod_{w \in \mathbf{d}_u} \prod_{j=1}^{N^{w}_u} (n_{z,-u}^{w} + \mathbf{\beta} +j - 1) }{\prod_{i=1}^{N_u}(n_{z,-u} + V \beta + i - 1)} \right\}
\end{split}
\end{equation}

\subsection{Choosing A New Cluster $P_{K+1}$:} \label{appdx: pnew}

To choose a new cluster, the number of clusters increases to $K+1$. Similarly, the corresponding probability of passenger $u$ choosing the new cluster is:

\begin{equation} \label{pnew} \tag{11}
\begin{split}
    &P(z_u = K+1 | \mathbf{z}_{-u}, \mathcal{D}, \mathbf{\alpha}, \mathbf{\beta}^O, \mathbf{\beta}^D, \mathbf{\beta}^T) \\
    & \propto P(z_u = K+1 | \mathbf{z}_{-u}, \mathcal{D}_{-u}, \mathbf{\alpha},  \beta^O, \beta^D, \beta^T)  \times P(\mathbf{d}_{u} | z_u = K+1, \mathbf{z}_{-u}, \mathcal{D}_{-u}, \mathbf{\alpha},  \beta^O, \beta^D, \beta^T)\\
    &\propto P(z_u = K+1 | \mathbf{z}_{-u}, \mathbf{\alpha}) \times  P(\mathbf{d}_{u} | z_u = K+1, \mathcal{D}_{z, -u},  \beta^O, \beta^D, \beta^T)
\end{split}
\end{equation}

We will derive the first term $ P(z_u = K+1 | \mathbf{z}_{-u}, \mathbf{\alpha}) $ and the second term $P(\mathbf{d}_{u} | z_u = K+1, \mathcal{D}_{z, -u},  \beta^O, \beta^D, \beta^T)$ with Appx. \ref{appdx: pnew_term1} and Appx. \ref{appdx: pnew_term2} respectively.

\subsubsection{The First Term of $P_{K+1}$}
\label{appdx: pnew_term1}

The first term is derived as follows:
\begin{equation} \label{pnew_term1} \tag{12}
\begin{split}
    &P(z_u = K+1 | \mathbf{z}_{-u}, \mathbf{\alpha}) = 1 - \sum_{k=1}^{K} P(z_u = k | \mathbf{z}_{-u}, \mathbf{\alpha}) =  1 - \frac{\sum_{k=1}^{K} m_{k, -u}}{M-1+\alpha}   = 1- \frac{M-1}{M-1+\alpha}   = \frac{\alpha}{M-1+\alpha}
\end{split}
\end{equation}

\subsubsection{The Second Term of $P_{K+1}$} \label{appdx: pnew_term2}

The second term is derived as follows:
\begin{equation}\label{pnew_term2}\tag{13}
\begin{split}
    & P(\mathbf{d}_{u} | z_u = K+1, \mathcal{D}_{z, -u},  \beta^O, \beta^D, \beta^T) \\
    & = \int P( \mathbf{d}_{u}, \phi_{K+1}^O, \phi_{K+1}^D, \phi_{K+1}^T | z_d = K+1, \beta^O, \beta^D, \beta^T ) d\phi_{K+1}^O d\phi_{K+1}^D d\phi_{K+1}^T \\
    & = \int [ P(\phi_{K+1}^O | z_u = K+1, \beta^O) P(\phi_{K+1}^D | z_u = K+1, \beta^D) P(\phi_{K+1}^T | z_u = K+1, \beta^T) \\
    & {\,\,\,\,\,\,\,\,\,\,\,\,}\times P( \mathbf{d}_{u} | {\,}\phi_{K+1}^O, \phi_{K+1}^D, \phi_{K+1}^T, z_u = K+1, \beta^O, \beta^D, \beta^T) ] {\,\,} d\phi_{K+1}^O d\phi_{K+1}^D d\phi_{K+1}^T\\
    & = \int [ \text{Dir}(\phi_{K+1}^O | \beta^O) \text{ Dir}(\phi_{K+1}^D | \beta^D) \text{ Dir}(\phi_{K+1}^T | \beta^T) \\
    & {\,\,\,\,\,\,\,\,\,\,\,\,}\times \prod_{i \in N_u}\text{Multi}(w_i^O | \phi_{K+1}^O) \text{ Multi}(w_i^D | \phi_{K+1}^D) \text{ Multi}(w_i^T | \phi_{K+1}^O) ] d\phi_{K+1}^O d\phi_{K+1}^D d\phi_{K+1}^T\\
    & = \int \frac{1}{\Delta( \beta^O )} \frac{1}{\Delta( \beta^D )} \frac{1}{\Delta( \beta^T )} \prod_{w^O} \phi_{K+1, w^O}^{\beta^O-1} \prod_{w^D} \phi_{K+1, w^D}^{\beta^D-1} \prod_{w^T} \phi_{K+1, w^T}^{\beta^T-1} \\
    & {\,\,\,\,\,\,\,\,\,\,\,\,} \times \prod_{w^O} \phi_{K+1, w^O}^{n_z^{w^O}-1} \prod_{w^D} \phi_{K+1, w^D}^{n_z^{w^D}-1} \prod_{w^T} \phi_{K+1, w^T}^{n_z^{w^T}-1} d\phi_{K+1}^O d\phi_{K+1}^D d\phi_{K+1}^T\\
    & = \frac{ \Delta( \mathbf{n}_{K+1} + \beta^O) }{ \Delta(\beta^O) }  \frac{ \Delta( \mathbf{n}_{K+1} + \beta^D) }{ \Delta(\beta^D) }  \frac{ \Delta( \mathbf{n}_{K+1} + \beta^T) }{ \Delta(\beta^T) } \\
    & = \frac{ \prod_{w^O} \Gamma(n_{K+1}^{w^O} + \beta^O) }{ \Gamma( \sum_{w^O} (n_{K+1}^{w^O} + \beta^O)) } \frac{ \Gamma( \sum_{w^O} \beta^O) }{ \prod_{w^O} \Gamma(\beta^O) } \times \frac{ \prod_{w^D} \Gamma(n_{K+1}^{w^D} + \beta^D) }{ \Gamma( \sum_{w^D} (n_{K+1}^{w^D} + \beta^D)) } \frac{ \Gamma( \sum_{w^D} \beta^D) }{ \prod_{w^D} \Gamma(\beta^D) } \times \frac{ \prod_{w^T} \Gamma(n_{K+1}^{w^T} + \beta^T) }{ \Gamma( \sum_{w^T} (n_{K+1}^{w^T} + \beta^T)) } \frac{ \Gamma( \sum_{w^T} \beta^T) }{ \prod_{w^T} \Gamma(\beta^T) } \\
    & = \frac{\prod_{w^O} \prod_{j=1}^{N_u^{w^O}} (\beta^O + j -1)}{\prod_{i=1}^{N_u} (V^O \beta^O + i - 1)} \times \frac{\prod_{w^D} \prod_{j=1}^{N_u^{w^D}} (\beta^D + j -1)}{\prod_{i=1}^{N_u} (V^D \beta^D + i - 1)} \times \frac{\prod_{w^T} \prod_{j=1}^{N_u^{w^T}} (\beta^T + j -1)}{\prod_{i=1}^{N_u} (V^T \beta^T + i - 1)} \\
    & = \prod_{ODT} \left\{ \frac{\prod_{w \in \mathbf{d}_u} \prod_{j=1}^{N_u^{w}} (\beta + j -1)}{\prod_{i=1}^{N_u} (V \beta + i - 1)} \right\}
\end{split}
\end{equation}

We substitute Eq. (\ref{pnew_term1}) and (\ref{pnew_term2}) into Eq. (\ref{pnew}), so that we could obtain the final result for $P_{K+1}$ as follows:

\begin{equation} \label{pnew_final} \tag{14}
\begin{split}
    P_{K+1} & = P(z_u = K+1 | \mathbf{z}_{-u}, \mathcal{D}, \mathbf{\alpha}, \mathbf{\beta}^O, \mathbf{\beta}^D, \mathbf{\beta}^T)\\
    &\propto P(z_u = K+1 | \mathbf{z}_{-u}, \mathcal{D}_{-u}, \mathbf{\alpha},  \beta^O, \beta^D, \beta^T) \times P(\mathbf{d}_{u} | z_u = K+1, \mathbf{z}_{-u}, \mathcal{D}_{-u}, \mathbf{\alpha},  \beta^O, \beta^D, \beta^T)\\
    & \propto \frac{\alpha}{M-1+\alpha} \prod_{ODT} \left\{ \frac{\prod_{w \in \mathbf{d}_u} \prod_{j=1}^{N_u^{w}} (\beta + j -1)}{\prod_{i=1}^{N_u} (V \beta + i - 1)} \right\}
\end{split}
\end{equation}

\section{Appendix: Tensor Gibbs Sampling for Tensor-DMM}
\label{app: tensor-gs-dmm}
This appendix introduce the Tensor Collapsed Gibbs Sampling algorithm for Tensor-DMM in detail.

\begin{algorithm}[H]
\SetAlgoVlined
\textbf{Input}: $\{\mathbf{d}_u\}_{u=1}^{M}, \alpha, {\beta}^{O}, {\beta}^{D}, {\beta}^{T}, max\_iter$\\
\textbf{Output}: $K, z_u, \forall u$\\
\textbf{Phase 1: Initialization}:\\
$K = K_0 (\underline{\text{ large enough}}), m_z, n_z, n_z^{w^O}, n_z^{w^D}, n_z^{w^T} =0$\\
\For{$u = 1$ to $M$}{
    Sample a cluster for $u$: $z_u \sim \text{Multi}(\frac{1}{K})$,
    $m_z = m_z + 1, n_z = n_z + N_u$\\
    \For{$i = 1$ to $N_u$}{
        $n_z^{w^O} = n_z^{w^O} + N_u^{w^O}$, same for D,T}
    }
\textbf{Phase 2: Gibbs Sampling:}\\
\For{$iter = 1$ to $max\_iter$}{
    \For{$u = 1$ to $M$}{
        \textbf{Phase 2.1: ``Kick-out'': }\\
        $m_z = m_z - 1, n_z = n_z - N_u$\\
        \For{$i = 1$ to $N_u$}{
            $n_z^{w^O} = n_z^{w^O} - N_u^{w^O}$, same for D,T
        }
        \If{$n_z ==0$}{
            $K=K-1$
        }
        \textbf{Phase 2.2: ``Choose A Cluster'': }\\
        $z_u \sim \{P_z\}_{z=1}^{K}$ in Eq. (\ref{eq_dmm_prob_choosek}) among $K$ clusters\\
        \textbf{Phase 2.3: ``Merge to the Selected Cluster'': }\\
        $m_z = m_z + 1, n_z = n_z + N_u$\\
        \For{$i = 1$ to $N_u$}{
            $n_z^{w^O} = n_z^{w^O} + N_u^{w^O}$, same for D,T
        }
    }
}
\caption{Tensor Collapsed Gibbs Sampling for Tensor-DMM}
\label{algo2}
\end{algorithm}

\printcredits

\bibliographystyle{cas-model2-names}

\bibliography{trc-updated}





\end{document}